\documentclass{article}

\usepackage{arxiv}
\usepackage[utf8]{inputenc} 
\usepackage[T1]{fontenc}    
\usepackage{hyperref}       
\usepackage{url}            
\usepackage{booktabs}       
\usepackage{amsfonts}       
\usepackage{nicefrac}       
\usepackage{microtype}      
\usepackage{doi}
\usepackage{lineno,hyperref,amssymb, multirow, slashbox, float, graphicx, tabu, amsmath}
\usepackage{natbib}

\title{Appearance based Deep Domain Adaptation \\ for the Classification of Aerial Images}

\author{ \href{https://orcid.org/0000-0002-0889-4171}{\includegraphics[scale=0.06]{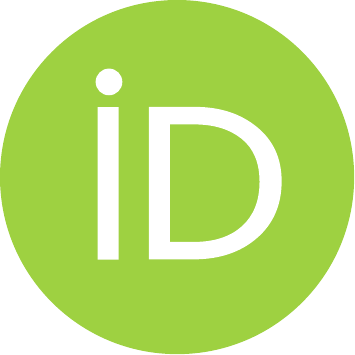}\hspace{1mm}Dennis~Wittich}\thanks{Corresponding Author} \\
	Institute of Photogrammetry and GeoInformation\\
	Leibniz University Hannover\\
	Germany \\
	\texttt{wittich@ipi.uni-hannover.de} \\
	\And
	\href{https://orcid.org/0000-0003-1942-8210}{\includegraphics[scale=0.06]{orcid.pdf}\hspace{1mm}Franz~Rottensteiner} \\
	Institute of Photogrammetry and GeoInformation\\
	Leibniz University Hannover\\
	Germany \\
	\texttt{rottensteiner@ipi.uni-hannover.de} \\
}

\date{March 2021}
\renewcommand{\undertitle}{Preprint submitted to ISPRS Journal of Photogrammetry and Remote Sensing}
\renewcommand{\headeright}{Preprint}

\begin{document}

\maketitle

\begin{abstract}

This paper addresses domain adaptation for the pixel-wise classification of remotely sensed data using deep neural networks (DNN) as a strategy to reduce the requirements of DNN with respect to the availability of training data. 
We focus on the setting in which labelled data are only available in a source domain $\mathcal{D}^S$, but not in a target domain $\mathcal{D}^T$. 
Our method is based on adversarial training of an appearance adaptation network (AAN) that transforms images from $\mathcal{D}^S$ such that they look like images from $\mathcal{D}^T$.
Together with the original label maps from $\mathcal{D}^S$, the transformed images are used  to adapt a DNN to $\mathcal{D}^T$. 
We propose a joint training strategy of the AAN and the classifier, which constrains the AAN to transform the images such that they are correctly classified.
In this way, objects of a certain class are changed such that they resemble objects of the same class in $\mathcal{D}^T$. 
To further improve the adaptation performance, we propose a new regularization loss for the discriminator network used in domain adversarial training.
We also address the problem of finding the optimal values of the trained network parameters, proposing an unsupervised entropy based parameter selection criterion which compensates for the fact that there is no validation set in $\mathcal{D}^T$ that could be monitored.
As a minor contribution, we present a new weighting strategy for the cross-entropy loss, addressing the problem of imbalanced class distributions.
Our method is evaluated in 42 adaptation scenarios using datasets from 7 cities, all consisting of high-resolution digital orthophotos and height data. 
It achieves a positive transfer in all cases, and on average it improves the performance in the target domain by $4.3\,\%$ in overall accuracy. 
In adaptation scenarios between the Vaihingen and Potsdam datasets from the ISPRS semantic labelling benchmark our method outperforms those from recent publications by $10-20\,\%$ with respect to the mean intersection over union. 


\end{abstract}



\section{Introduction}
\label{sec:Introduction}


The task of pixel-wise image classification is to assign a class label to each pixel in the image according to a pre-defined class structure. 
In remote sensing (RS) applications, the data usually consist of an orthorectified multi-spectral image (MSI), and, possibly, height information, e.g. obtained by 3D-reconstruction from overlapping images. 
For several years, research on this topic has been dominated by supervised methods based on deep learning, in particular on Fully Convolutional Neural Networks (FCN) \citep{Long_2015_CVPR}, e.g. \citep{marmanis2016semantic, zhang2019dual}. 
One of the main problems related to deep learning in RS is the requirement of FCN for the availability of a large amount of training samples, the generation of which involves a labour-intensive interactive labelling process \citep{Zhu_2017_DL_RS_Zfs}. 
One strategy to solve this problem is {\em domain adaptation} ({\em DA}) \citep{tuia2016}, a special setting of  transfer learning (TL) \citep{pan2010}. 
This paper presents a new method for DA dedicated to the pixel-wise classification of RS data based on deep learning. 


In DA, data are assumed to be available in different domains. 
The domains share the same feature space, but the data might follow different distributions. 
In both domains, a learning task is to be solved, characterized by the same class structure. 
One distinguishes a source domain $\mathcal{D}^S$, in which an abundance of training samples with known class labels is available, and a target domain  $\mathcal{D}^T$. 
In {\em semi-supervised DA}, the scenario we are interested in, no labelled samples are available in $\mathcal{D}^T$  \citep{tuia2016}.
The main goal of DA is to use the information available in $\mathcal{D}^S$ to find a better solution for the task in $\mathcal{D}^T$, which requires the domains to be related \citep{pan2010}. 


In RS applications, the domains can be associated with imagery from different geographical regions or acquired at different epochs, but having the same input channels. 
The source domain corresponds to images for which pixel-level class annotations are known, e.g. from earlier projects, whereas the target domain corresponds to a new set of images to be classified according to the same class structure. 
In semi-supervised DA (which we will simply refer to  as {\em DA} in the remainder of this paper) this is to be achieved without having to generate (new) training labels in $\mathcal{D}^T$, even though there is a {\em domain gap}, i.e. even though the distributions of the data in the two domains are different \citep{tuia2016}. 
Due to this domain gap, a classifier trained using the labelled data from $\mathcal{D}^S$ will achieve a lower performance in $\mathcal{D}^T$ compared to a classifier trained using labelled data from $\mathcal{D}^T$.
The goal of DA is to reduce this performance gap, i.e. to achieve a {\em positive transfer}, while at the same time avoiding a {\em negative transfer}, i.e. a reduction in the classifier's performance after DA.


There are different strategies for DA. 
Methods based on \textit{instance transfer} start with training a classifier using data from $\mathcal{D}^S$. 
This classifier is applied to data from $\mathcal{D}^T$ to predict {\em semi-labels}, and the classifier is  re-trained using target domain data with semi-labels in an iterative procedure. 
We believe that this strategy is strongly limited by the initial performance of the (source) classifier in $\mathcal{D}^T$.
The second strategy for DA, used frequently in the context of deep learning \citep{wang2018deep}, is based on \textit{representation transfer}. 
Such methods map images from both domains to a shared and domain-invariant representation space in which a classifier trained using samples from $\mathcal{D}^S$ can also be used to classify data from $\mathcal{D}^T$, e.g.  \citep{tzeng2017,liu19curve}. 
However, we found such an approach to be rather difficult to train and to be susceptible to result in a negative transfer \citep{wittich19adda}.


Consequently, the approach presented in this paper follows another strategy, which we refer to as {\em appearance adaptation} and which is based on methods for style transfer \citep{zhu2017unpaired, liu2017}. 
It can be seen as a special case of representation transfer in which the common representation is the space of the original radiometric features of $\mathcal{D}^T$. 
A common approach is to use unlabelled samples from both domains to train an {\em appearance adaptation network} that transforms images from $\mathcal{D}^S$ so that they look like images from $\mathcal{D}^T$. 
As the label information is not changed in this process, the transformed source images with known labels are used to train a classifier for $\mathcal{D}^T$. 
This strategy was originally applied to street scene segmentation \citep{hoffman2018cycada, zhang2018fully}, examples for its application in RS are \citep{benjdira2019unsupervised,tasar2020colormapgan,tasar2020semI2I,li2020GAN4RS,soto2020domain, GriOst2020}. 


A major problem of appearance adaptation is to achieve what we call \textit{semantic consistency}: it is not sufficient for the transformed images to give an overall impression similar to the images from $\mathcal{D}^T$, but the appearance of objects has to be adapted so that after the transformation the objects of any given class look similar to the objects of the same class in $\mathcal{D}^T$ (e.g., pixels corresponding to buildings should look like buildings of $\mathcal{D}^T$ after the adaptation).
This is particularly difficult if the label distributions in $\mathcal{D}^S$ and $\mathcal{D}^T$ are very different \citep{soto2020domain, GriOst2020}. 
\citet{soto2020domain} tried to solve this problem by applying a {\em cycle consistency} constraint \citep{zhu2017unpaired}, in which images transformed from  $\mathcal{D}^S$ to  $\mathcal{D}^T$ and back again using a second adaptation network are required to have the same grey values as the original ones. 
This approach resulted in artifacts in the adapted images, a problem called {\em hallucination of features} \citep{cohen2018hallus}. 
\citet{GriOst2020} tried to align the label distributions based on label maps predicted in $\mathcal{D}^T$, but this did not lead to significant improvements in DA. 
Approaches trying to constrain the label distribution in  $\mathcal{D}^T$ to be similar to the one in $\mathcal{D}^S$, e.g. \citep{zhangetal2017}, may even be detrimental in RS, where real differences in the label distributions are available and have been shown to be problematic for DA \citep{wittich19adda}. 
We conclude that achieving semantic consistency remains an unsolved problem in appearance-based DA, especially in presence of large differences in the label distributions in $\mathcal{D}^S$ and $\mathcal{D}^T$.


Another unsolved problem in DA is \textit{parameter selection}, i.e. the selection of the values of the network parameters to be used for classification, which can also be seen as the problem of selecting a stopping criterion for training.
\citet{GriOst2020} report that the classification error in $\mathcal{D}^T$ shows large variations in the DA process and may even increase again after reaching a minimum, a behaviour we also observed in \citep{wittich19adda, wittich20entmin}.
In classical machine learning, a validation dataset is used to select a proper epoch to stop the training process. 
However, in our DA scenario, there are no labelled samples in $\mathcal{D}^T$ and, thus, there is no validation set. 
In such a context, parameter selection is usually solved by making the number of training epochs a hyper-parameter to be tuned and using the parameters values determined in the final iteration.
However, if one DA scenario with labels in $\mathcal{D}^T$ is used to tune this hyper-parameter, there is no guarantee that the optimal number of training epochs in DA can be transferred to other pairs of source and target domains. 
It would be desirable to be able to select the appropriate epoch for stopping the training process based on some other criterion. 


In this work we address these unsolved problems by proposing a new method for DA based on appearance adaptation.
It applies adversarial training \citep{goodfellow2014generative}, in which a discriminator network learns to distinguish adapted images from real target images, while the adaptation network learns to fool the discriminator by transforming the images such that they look like images from $\mathcal{D}^T$ \citep{zhu2017unpaired}. 
However, unlike most existing approaches we do not apply cycle-consistency to constrain the adaptation, but try to achieve semantic consistency by {\em joint training of the appearance adaptation and classification networks}. 
Thus, the appearance adaptation network does not only learn to transform images from $\mathcal{D}^S$ so that they look like images from $\mathcal{D}^T$, but the transformed images still need to be classified correctly, which we believe to be a suitable measure to avoid that transformed regions belonging to a certain class will look like areas corresponding to another class in $\mathcal{D}^T$. 
The contributions of this paper can be summarized as follows:


\begin{enumerate}
	\item We present a novel approach for DA based on semantically consistent appearance adaptation that relies on adversarial training of an appearance adaptation network jointly with the classification network. The approach requires only a single adaptation network from $\mathcal{D}^S$ to $\mathcal{D}^T$, which makes it less memory consumptive and easier to tune than existing methods. 
	\item To further improve semantic consistency we introduce a new regularization term to adversarial training to mitigate problems due to large differences in the class distributions in $\mathcal{D}^S$ and $\mathcal{D}^T$. It prevents the discriminator from learning trivial solutions for differentiating samples from different domains.
	\item We propose a new criterion for selecting the optimal parameter values in DA that does not require any labelled validation data in $\mathcal{D}^T$. 	It relies on an entropy-based confidence measure for the predictions in $\mathcal{D}^T$. 
	\item As a minor contribution, we address the problem of a poor classification performance for classes that are under-represented in the training set, presenting a new adaptive cross-entropy loss function which uses class-wise performance metrics to tune the weights of samples in supervised training.
\end{enumerate}



\section{Related Work}
\label{sec:relwork}




\subsection{Semi-supervised deep domain adaptation}
\label{sec:relwork-DA}


This review 
discusses methods for adversarial representation transfer, appearance adaptation, and hybrid methods combining different strategies.  
A more generic overview on deep DA can be found in \citep{wang2018deep}.

\subsubsection{Adversarial methods for representation transfer}
\label{sec:relwork-DA-RT}


The main idea of representation transfer is to map images from both domains to a shared representation space in which a single classifier is applied.
Many approaches for representation transfer rely on adversarial training.
Originally developed for methods designed to predict a single class label for every input image  \citep{ganin2016domain,tzeng2017adversarial}, this approach has also been transferred to pixel-wise classification of street scenes \citep{huang2018domain}. 
In this application, an important motivation for DA is the desire to use synthetic images to generate training samples with pixel-wise label annotations for training (source domain $\mathcal{D}^S$) and adapt the resultant classifier to real (target) images. 
However, we think that the success of DA in this scenario is strongly related to the fact that typical street scenes are rather similar to each other with respect to the class distributions in the two domains. 
In some approaches, this is used to constrain the DA, e.g. to deliver a label distribution in $\mathcal{D}^T$ similar to the one in $\mathcal{D}^S$, e.g. \citep{zhangetal2017, tuan2019advent}, or to consider prior information about the typical location of objects in a street scene (e.g., sky is expected to occur in the upper part of an image)  \citep{you2018cbst}. 
However, for reasons already discussed, such assumptions are not justified in RS applications.


An example for DA based on representation transfer in RS is \citep{riz2016_autoencoder}. 
A domain-independent feature representation from images of two geographical areas is obtained by training a stacked auto-encoder using images from both domains that learns to reconstruct the input image via a lower-dimensional feature space. 
This seems to work well for domains that are rather similar, but it remains unclear whether it would still be sufficient in the presence of larger domain differences.
\citet{GriOst2020} perform representation transfer based on a domain distance for the pixel-wise classification of aerial images. 
Their results show that the adaptation performance strongly decreases if the class distributions are very different in the two domains. 
To improve the adaptation performance they align the representations using target images found to be semantically similar according to label maps predicted by the classifier trained on source domain data, but this only leads to an improvement in a half of the presented experiments.
\citet{liu19curve} aim at representation transfer by matching so-called feature curves from both domains using adversarial training. 
However, the domain gap that could be bridged by this method was limited. 
This indicates that adversarial representation transfer is difficult if the domain gap is large, e.g. when adapting between two different cities in which the objects have a different appearance or in which the class distributions are dissimilar, both of which is the case for the public benchmark dataset used in \citep{liu19curve}.
We use this approach as a baseline for comparison in our experiments. 
In \citep{wittich19adda} we also used representation transfer based on adversarial training for DA.  
We could achieve a small but stable improvement of the classification results due to DA if an early network layer was chosen for transfer. 
However, the results strongly depended on the hyper-parameters used in training, which makes this approach difficult to tune. 


From the overview presented in this section we conclude that approaches based on representation transfer are facing problems in the presence of a large domain gap, in particular in case of large differences in the class distributions of the two domains. 
Approaches for stabilizing adversarial representation transfer for street scene classification frequently rely on assumptions which are not generally justified in RS applications, e.g. on class distributions to be similar.


\subsubsection{DA based on appearance adaptation}
\label{sec:relwork-DA-AA}


This group of methods can be seen as a special case of representation transfer in which the shared representation space is the original feature space of the images in one domain.
Relying on concepts for style transfer, they take an image from $\mathcal{D}^S$ and adapt it so that it looks as if it were a sample from $\mathcal{D}^T$.
As the training labels from $\mathcal{D}^S$ are not affected by this transformation, a classifier can be trained in a supervised way based on the transformed images, e.g. \citep{yang2020phase, yang2020fda, chang2019all, musto2020semantically, chen2019crdoco, hoffman2018cycada}. 
Maintaining \textit{semantic consistency} as defined in section~\ref{sec:Introduction} is crucial for the success of this strategy.
Some authors try to achieve this goal in the frequency domain, where the adaptation is applied to the amplitude, either based on learning \citep{yang2020phase} or by swapping the low frequency coefficients between the source and target images \citep{yang2020fda}.
We believe that in RS it may sometimes also be required to consider modifications of higher frequencies,
e.g. when transferring between domains corresponding to images acquired at different seasons, in which deciduous trees look completely different.


In the field of RS, \citet{benjdira2019unsupervised} used CycleGAN \citep{zhu2017unpaired} to transform images from two different cities which were treated as different domains. 
Their main goal is to learn semantically consistent transformations between both domains by incorporating a cycle-consistency loss. 
The method, which we use as another baseline to compare our method to, results in quite large improvements in the classification performance for two out of six classes due to DA, but the results for the other classes could barely be improved. 


Other papers on DA for RS applications present modifications of CycleGAN designed to improve the DA performance.
\citet{soto2020domain} address deforestation detection based on satellite imagery, defining the domains to correspond to images of different geographical regions.
They show that cycle consistency is not sufficient to preserve the semantic structure in the adaptation process, apparently because the differences in the class distributions of the domains lead to hallucinated structures in the transformed images.
They introduce an additional identity mapping loss which reduces the amount of hallucinated structures, but also leads to a decreased performance of the classifier after DA.
\citet{GriOst2020} address the pixel-wise classification of urban scenes. 
Observing plain CycleGAN to lead to a negative transfer in $50\,\%$ of their experiments, the authors tried to improve the adaptation by training on semantically paired images (cf.~section \ref{sec:relwork-DA-RT}), but this did not result in a significant improvement.


There are also strategies to achieve semantic consistency that do not require cycle-consistency. 
\citet{tasar2020colormapgan} learn a colour mapping to perform the image adaptation from the source to the target domain. 
However, as this approach cannot adapt the texture of objects, we think it is too limited to work in more complex DA scenarios, e.g. requiring transfer between images from different seasons.
\citet{tasar2020semI2I} use a bi-directional image-to-image transformation based on an alternative to cycle-consistency called cross-cycle-consistency \citep{lee2018diverse} and an alignment of  the image gradients between the images before and after the transformation to achieve semantic consistency. 
In our opinion, this may be too strong a regularization when trying to apply DA to imagery from different seasons, because the gradient maps may change a lot in vegetated areas.


The method proposed in this paper is also based on appearance adaptation. 
However, compared to the papers cited in this section we propose a different strategy to achieve semantic consistency. 
Instead of relying on cycle-consistency or cross-cycle-consistency, we only train a single adaptation network that transforms images from the source to the target domain.
Semantic consistency is achieved by not only enforcing the transformed source images to look like those from $\mathcal{D}^S$ after adaptation, but also by requiring them to be classified correctly after the transformation.
We also address possible problems caused by different class distributions in the two domains by applying a new  regularization term to the output of the discriminator network. 
To the best of our knowledge, ours is the first approach for appearance adaptation based on a joint training the adaptation and the classification networks in the context of RS.


\subsubsection{DA based on hybrid approaches}
\label{sec:relwork-DA-HA}


Quite a few hybrid approaches combine representation matching and appearance adaptation.
For instance, \citet{zhang2018fully} use gradient-based style transfer \citep{gatys2016image} to reduce the visual difference between the two domains before feeding the images to the classification network, where representation transfer is applied. 
However, this approach leads to high computation times and requires a considerable amount of hyper-parameter tuning to achieve good results.
One of the first hybrid approaches combining CycleGAN for appearance adaptation and adversarial representation transfer was CyCADA \citep{hoffman2018cycada}, applied to street scene classification.  
It is based on a rather complex network which, according to the authors, cannot be trained end-to-end on a consumer GPU due to very high memory requirements. 
\citet{musto2020semantically} extend CyCADA by feeding the predicted label map to the appearance adaptation network and enforcing consistency between the label maps predicted for both, the original and the transformed source images. 
\citet{chen2019crdoco}  additionally enforce the predictions of original target images and adapted target images to be consistent, whereas \citet{chang2019all} replace cycle-consistency with cross-cycle-consistency.
All of these methods are very complex with respect to the number of modules, parameters and training procedures. 
This may be the reason why none of them directly uses the classification loss for the transformed source images jointly with the losses related to the adaptation network to enforce semantic consistency. 
In contrast, our method does not combine appearance adaptation with representation transfer, but only learns the appearance adaptation and the classifier, which reduces the network complexity. 


An architecture very similar to CyCADA was trained end-to-end in \citep{i2i4da}. 
The authors learn two encoders  which  embed images from both domains in a shared feature space by adversarial training. 
Simultaneously, two decoders are trained, which recover images based on the embeddings.
To enforce semantic consistency the authors use an identity loss that enforces recovered representations to look like the original inputs. 
Further, they learn image-to-image transformations by decoding representations from each domain to the corresponding other one.
The second aspect is again achieved via adversarial training, requiring two additional discriminators.
The actual classifier is optimized to correctly classify embeddings for the source domain and embeddings for images transformed from $\mathcal{D}^S$ to $\mathcal{D}^T$.
The approach achieves good results for the pixel-wise classification of street scenes. 
However, it is unclear if it is transferable to DA in RS, where different class distributions pose additional challenges \citep{wittich19adda}.
We consider \citep{i2i4da} to be the work closest to our approach, because it is also based on the joint training of the appearance adaptation and the classification networks. 
However, we do not map the images to some domain-independent intermediate representation, but only adapt images from $\mathcal{D}^S$ to $\mathcal{D}^T$. 
Thus, we only need  one domain discriminator and one appearance adaptation network, which largely reduces the overall memory footprint of the architecture. 
Secondly, we propose an additional regularization of the discriminator, addressing the problems due to large  differences in the class distributions. 


Whereas there are many hybrid methods for street scene classification, we found only one such approach addressing the pixel-wise classification of RS images.
\citet{ji2020fullspace} combine appearance adaptation and representation transfer, using adversarial training in both cases.  
For the appearance adaptation they also rely on cycle-consistency. 
Representation transfer is applied in the last layer of the network.
As already discussed in section \ref{sec:relwork-DA-AA}, we think that cycle consistency may not be sufficient if the domains are very different.
However, because the authors report quite good results on publicly available datasets, we use this approach as another baseline for the evaluation of our method.


\subsection{Stopping and parameter selection criteria}
\label{sec:relwork-PS}


A problem that is barely addressed in research on DA is the stopping criterion for the adaptation. 
In supervised training, it is common practice to monitor the performance of the classifier on a validation set and to select the parameter values resulting in the best validation performance as the final result of the training process \citep{prechelt1998early}.
However, this approach cannot be used for semi-supervised DA for lack of labelled samples that could be used for validation in the target domain.
Unfortunately, in DA it is particularly important for how long the adaptation process is continued. 
For instance, \citet{GriOst2020} show that after increasing for some time, the performance on a test set from the target domain decreases if the adaptation is carried out for too long.
\citet{benaim2018estimating} discuss this problem for unsupervised appearance adaptation.
However, they do not propose a solution, but derive a bound to predict the success of such methods.
The common strategy in DA is to fix the number of epochs for the adaptation and the very last parameter set is used for inference \citep{tasar2020semI2I, tasar2020colormapgan, benjdira2019unsupervised, musto2020semantically}, which means that this hyper-parameter has to be tuned with care. 
Some publication do not even tell for how many epochs they train their model \citep{i2i4da, liu19curve,hoffman2018cycada,chen2019crdoco}.
To the best of our knowledge, this paper proposes the first method to solve the problem of unsupervised parameter selection in DA. 


\subsection{Training with imbalanced data}
\label{sec:relwork-ID}


In RS applications, the class distribution of the training samples is often imbalanced.
In such a case, the cross-entropy loss, which is commonly used for training  FCNs, is dominated by frequent classes, so that after training, the prediction quality of under-represented classes may be not satisfactory. 
One way to compensate this imbalance is to take measures which lead to well-defined clusters in the feature space.
This can be achieved by considering similarity measures like the Euclidean distance \citep{hadsell2006, schroff2015facenet} or the cosine similarity \citep{yang2020} of latent representations of samples belonging to the same class as constraints in the loss function.
\citet{hadsell2006} and \citet{schroff2015facenet} have shown that a clustering approach based on the Euclidean distance of representations can improve the results in tasks related the prediction of a single label per image, but it remains unclear if this also applies for pixel-wise classification.
\citet{yang2020} could improve the mean F1-score in a scenario addressing the pixel-wise classification of aerial images by enforcing the cosine similarity of representations belonging to the same class to be close to the respective centroid.
However, we think that the cosine similarity does not necessarily lead to compact clusters, as it mainly affects the directions of vectors in feature space.  


Another way of compensating class imbalance is to use a weighted cross-entropy loss in which pixels which correspond to an under-represented class are considered with a higher weight than pixels of the more frequent classes.
However, this approach may be problematic in a DA scenario, because the class distribution in the target domain is unknown.
Alternative loss functions such as the focal loss for binary classification \citep{lin2017focal} or its variant for the multi-class case \citep{yang2019focal} define the weights according to the predicted score for the reference class of each sample. 
In this way, the training process should focus on pixels that were predicted with high uncertainty, which was shown to increase the prediction quality of under-represented classes to some extent. 
However, pixels with a low confidence frequently correspond to pixels at object boundaries, where the label information is uncertain due to geometrical inaccuracies. 
In this work, we propose a different approach which incorporates class-wise performance metrics, but does not do this for every sample individually. 
The basic idea is that classes which are predicted with a lower quality should have a higher impact on the overall loss.
An alternative  would be to use the dice loss \citep{sorensen1948dice, ren2020full}, but this would cause the classifier to focus too much on object borders. 
We think that this can be suboptimal for the reasons given in the context of the focal loss.



\section{Methodology}
\label{sec:methodology}


We start with a formalization of the task  following \citep{tuia2016}. 
Let  $x_k$ be an input image of size ${H\times W \times N}$ from the input feature space $\mathcal{X} = \mathbb{R}^{H\times W \times N}$. 
$H$ and $W$  denote the height and width of the image, respectively, and $N$ is the number of channels.
If height information is available, the first $N-1$ channels correspond to the orthorectified multispectral image and the last channel contains the metric height information for each pixel, e.g. in the form of a normalized digital surface model (nDSM) which contains the height above terrain for every pixel. 
Each input image $x_k$ corresponds to a label map $y_k$ that encodes the semantic label of every pixel. 
Thus, $y_k$ comes from the categorical label space $\mathcal{Y} = \mathbb{L}^{H \times W }$ where $\mathbb{L} = \{1,..2,..,l\}$ and $l$ is the number of classes in the pre-defined class structure. 
The learning task is to find a parameter set $\Theta_C$ of a classifier $C(\cdot)$ such that it predicts the correct label map $y_k$ for any input image $x_k$. 
In this work, we consider two domains, a source domain $\mathcal{D}^S$, where a training set $T^S = \{x_k^S, y_k^S\}_{k=1}^n$ of $n$ labelled images is assumed to be available, and a different but related target domain $\mathcal{D}^T$, where only the set $U^T = \{ x_k^T\}_{k=1}^m$ of $m$ unlabelled images is available. 
We address the semi-supervised setting of domain adaptation, trying to use $T^S$ and $U^T$ jointly to solve the learning task in $\mathcal{D}^T$ such that the resulting model achieves a better performance than a classifier trained only on $T^S$. 
This corresponds to a common scenario in RS, were a dataset labelled in the past is used for training a classifier which is to be applied to new data without requiring new training labels.


\subsection{Overview}
\label{sec:methodology:overview}

\begin{figure}[!ht]
	\centering
	\includegraphics[width=1\textwidth]{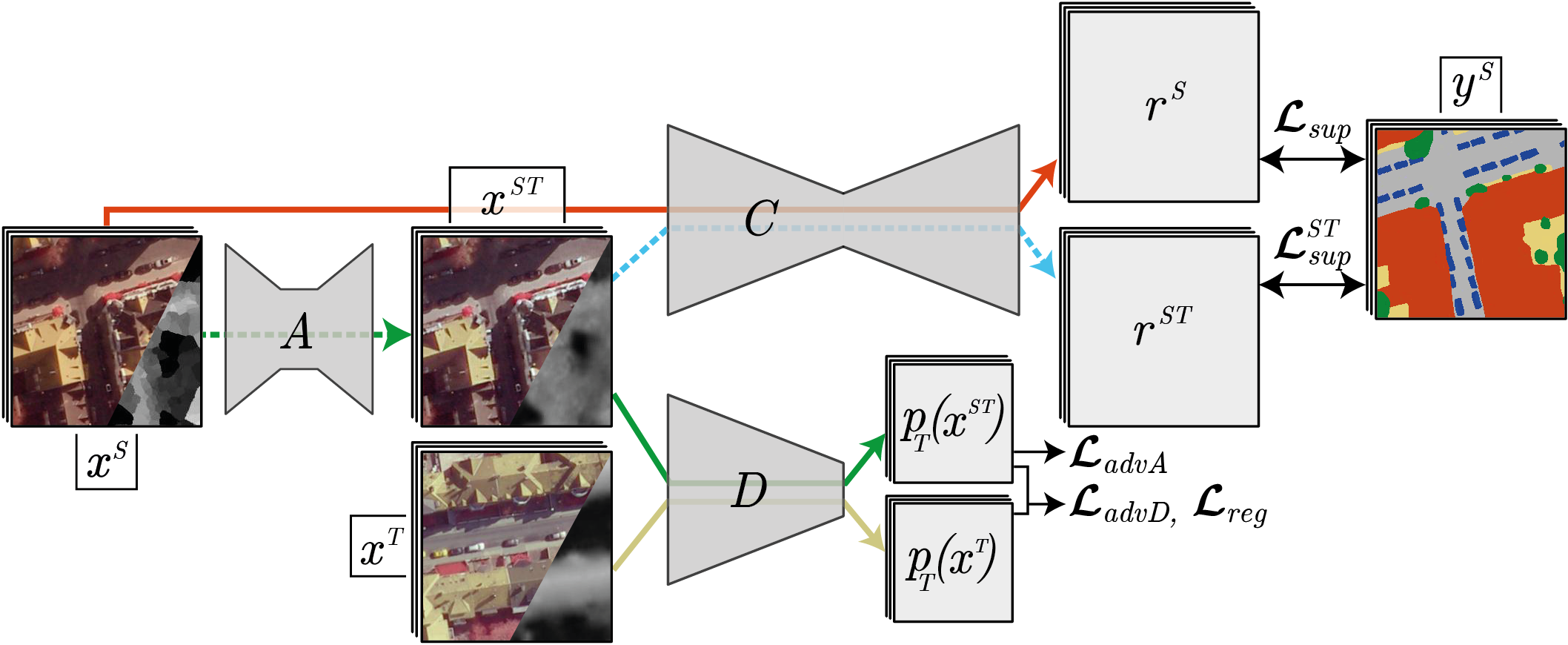}
	\caption{
		Method overview. 
		The classification network \textit{C} predicts maps of class scores $r^S$ and $r^{ST}$ for source images ($x^S$) and transformed images ($x^{ST}$), respectively, the latter being produced by the appearance adaptation network \textit{A}. 
		In both cases, \textit{C} is trained to predict label maps that match the reference $y^S$ from $\mathcal{D}^S$ by minimizing the supervised loss terms $\mathcal{L}_{sup}$ and $\mathcal{L}_{sup}^{ST}$. Simultaneously, \textit{A} is trained in an adversarial way using the discriminator \textit{D}. 
		To enforce semantic consistency, \textit{A} is also trained to minimize $\mathcal{L}_{sup}^{ST}$. \textit{D} delivers probability maps $p_{T}(\cdot)$ for the corresponding inputs to belong to $\mathcal{D}^T$. 
		They are considered in the adversarial loss terms $\mathcal{L}_{advA}$ and $\mathcal{L}_{advD}$ and in the proposed regularization term $\mathcal{L}_{reg}$. 
	}\label{fig:overview}
\end{figure}

An overview of the method including the main loss terms used in training is shown in figure~\ref{fig:overview}. 
As in all DA methods based on appearance adaptation, the core idea is to substitute missing label information in $\mathcal{D}^T$ by labelled images from $\mathcal{D}^S$ that were transformed such that they have an appearance similar to images from $\mathcal{D}^T$. 
This is achieved by an appearance adaptation network \textit{A}.
Denoting the transformed version of a source image $x^S_k$ by $x^{ST}_k$, these transformed images are used jointly with the corresponding label maps $y_k$ from the source domain to train a classifier $C$ in a supervised way. 
By passing the transformed images through the network and optimizing its parameters such that the predictions are correct, the classification network should be adapted to the target domain. 
For this approach to be successful, the transformed images should look like images from the target domain, but the transformation also has to be semantically consistent in the way defined in section~\ref{sec:Introduction}.
To that end, we propose a joint training of \textit{A} and \textit{C}, using the supervised loss of the transformed images $\mathcal{L}^{ST}_{sup}$ as a guidance for \textit{A} to achieve semantic consistency. 
To make the transformed images look like images from the target domain, we rely on adversarial training of \textit{A} and a domain discriminator \textit{D}.
The idea is that \textit{A} learns to transform input images from $\mathcal{D}^S$ to $\mathcal{D}^T$ such that they look like coming from $\mathcal{D}^T$ (adversarial training), yet still being correctly classified by \textit{C} (supervised guidance). 
The latter aspect is the main key to achieve a good performance in the target domain, but simultaneously, due to another supervised loss $\mathcal{L}_{sup}$, \textit{C} also learns to classify images from the source domain, which is required to achieve semantic consistency (cf. \ref{sec:appearance_adaptation}).
In addition, we introduce a regularization loss $\mathcal{L}_{reg}$ which should prevent \textit{D} from learning to differentiate the domains only based on the occurrence of simple features and, thus, should prevent \textit{A} from hallucinating structures. 
In the subsequent sections the network architecture and the training strategy are described in detail.


\subsection{Network architecture} 
\label{sec:methodology:network}


The network architecture consists of the three modules: the classification network \textit{C}, the appearance adaptation network \textit{A} and the domain discriminator \textit{D} with corresponding sets $\Theta_A$, $\Theta_C$ and $\Theta_D$ of trainable parameters. 


\subsubsection{Classification network}
\label{sec:methodology:network:classinet}


\begin{table}[!b]
	\begin{center}
		\begin{tabular}{c l  c  c  c r}
			&Layer & Layer type & H / W      &         Depth    &\\
			\hline
			\hline
			\multirow{9}{*}{\rotatebox[origin=c]{90}{\parbox[c]{2.7cm}{\centering Encoder}}}
			&1 & Input layer                &  256 &   $N$ & \\
			&2 & Conv(3) stride 2, BN, ReLU &  128 &    32 & \\
			&3 & Conv(3), BN, ReLU          &  128 &    64 &*\\
			&4 & Xception block             &   64 &   128 &*\\
			&5 & Xception block             &   32 &   256 &*\\
			&6-15 & Xception block          &   16 &   728 &*\\
			&16 & Xception block            &    8 &  1024 &*\\
			&17 & Conv(3), BN, ReLU         &    8 &  1536 &*\\
			&18 & Conv(3), BN               &    8 &  2048 &*\\
			
			\hline
			\multirow{11}{*}{\rotatebox[origin=c]{90}{\parbox[c]{2.7cm}{\centering Decoder}}}
			&16 & Upsample, Concat(15)      &   16 & 2776 & \\
			&17, 18 & Conv(3), ReLU         &   16 &  256 & \\
			&19 & Upsample, Concat(5)       &   32 &  512 & \\
			&20, 21 & Conv(3), ReLU         &  32  &  128 & \\
			&22 & Upsample, Concat(4)      &  64  &  256 & \\
			&23, 24 & Conv(3), ReLU         &  64  &   64 & \\
			&25 & Upsample, Concat(3)       &  128 &  128 & \\
			&26, 27 & Conv(3), ReLU         &  128 &   32 & \\
			&28 & Upsample                  &  256 &   32 & \\
			&29, 30 & Conv(3), ReLU         &  256 &   16 & \\
			&31 & Conv(1), Softmax          &  256 &  $l$ & \\
		\end{tabular}
		\caption{
			\label{tab:segmentation}Classification network $C$. 
			Conv($s$): convolution with kernel size $s\times s$; BN: Batch-Normalization; ReLU: rectified linear unit. 
			Layers marked with an asterisk are pre-trained on the ImageNet dataset \citep{deng2009imagenet}. \textit{Concat(X)}:  depth-wise concatenation of the output of the layer \textit{X} and the current layer. H / W / Depth: output dimensions. }
	\end{center}
\end{table}


The classification is performed by a FCN which takes an image $x_k$ with $N$ channels as input and predicts pixel-wise class scores. 
These scores are normalized by applying the softmax function, which results in probabilistic class scores for each pixel, arranged in maps $r^S_k$ for images from $\mathcal{D}^S$ and $r^{ST}_k$ for transformed images $x^{ST}_k$. 
This corresponds to the red and blue paths in figure~\ref{fig:overview}, respectively.
The actual class predictions are obtained by selecting the class with the highest probability for each pixel.
In table \ref{tab:segmentation} the full architecture of the classification network is given. 
We use a UNet-like architecture \citep{unet} with an Xception backbone \citep{xception} pre-trained on ImageNet \citep{deng2009imagenet}. 
The Xception backbone corresponds to the encoder of our classification network, including all layers before the one with the lowest spatial resolution (layer 18 in table~\ref{tab:segmentation}).
In the decoder of the network, nearest neighbour interpolation is used for upsampling. 
In preliminary experiments we compared this architecture in combination with a pre-trained initialization to a residual network with completely random initialization, similarly to \citep{wittich20entmin}. 
We observed a comparable performance but a noticeable reduction of the training time.
Because the backbone is pre-trained on three-channel RGB-images, the first layer of the network cannot be used if the number $N$ of input channels is different from three.
In that case, the first layer is replaced by a convolution layer with $N$ input-channels which is initialized randomly. 
The parameters of the decoder are also initialized randomly; all random initializations are based on \citep{he2015delving}. 
Overall, this network has about 28.8 M parameters. 
We choose this rather large network because it has to learn to classify images from $\mathcal{D}^S$ and from $\mathcal{D}^T$ (cf. section \ref{sec:training}), which we assume to be a more complex task than classifying only images from one domain due to the higher variability of the data. 
By using a larger network, we reduce the risk that the learning capacity of \textit{C} becomes a limiting factor of the method.


\subsubsection{Appearance adaptation network}
\label{sec:methodology:network:appearance}


The appearance adaptation network takes an image $x_k^S$ from the source domain as input and delivers the transformed version $x_k^{ST}$. 
This corresponds to the dotted green line in figure~\ref{fig:overview}.
For this task we use a residual FCN with about 5 M parameters that is a simplified version of the one used in \citep{wittich20entmin}. 
Table \ref{tab:generator} lists all layers of the network. 
An initial convolution with a stride of four pixel downsamples the signal by a factor of 4. 
It is followed by 15 residual blocks at the reduced scale. 
Each of them consists of two subsequent $3\times 3$ convolutions with 64 and 256 filters, respectively, zero padding and ReLU activation.
The result of each residual block is added to its input. 
At the end, two transposed convolutions are used to enable predictions at the original spatial resolution of the input. 
Commonly, non-linearities with a bounded range of values such as the hyperbolic tangent function are used to produce the output of networks for image generation (regression of colour values) \citep{goodfellow2014generative}. 
However, because we also want to be able to predict transformed height information we do not restrict the values to a fixed range. 
Consequently, we do not apply any activation function to the output of the last convolutional layer.


We chose a residual network for the appearance adaptation because we expect the optimal solution for this task not to deviate too much from an identity mapping.  
For example, objects which have the same appearance in both domains should not be changed at all in the adaptation.
Following \citep{he2016}, we assume that residual networks are well suited to learn such a solution, which we confirmed in preliminary experiments. 


\begin{table}[!ht]
	\begin{center}
		\begin{tabular}{ l  c  c  c }
			Layer & Layer type& H / W  & Depth \\
			\hline
			\hline
			1 & Input layer                   &  256 & $N$ \\
			2 & Conv(6) stride 4, ReLU, BN    &  64  & 256 \\
			3-18 & Residual block             &  64  & 256 \\
			19 & T-Conv(4) stride 2, BN, ReLU &  128 & 128 \\
			20 & T-Conv(4) stride 2           &  256 & $N$ \\
		\end{tabular}
		\caption{\label{tab:generator}Residual network for appearance adaptation. T-Conv($s$): transposed convolution with kernel size $s\times s$. For other abbreviations, cf.~table~\ref{tab:segmentation}.}
	\end{center}
\end{table}


\subsubsection{Domain discriminator}
\label{sec:methodology:network:discriminator}


The discriminator network is required for the training of the appearance adaptation network in an adversarial way (cf.~section \ref{sec:appearance_adaptation}). 
Its task is to predict whether an image is from the target domain or whether it is a transformed source image.
However, instead of predicting one class score per image \citep{goodfellow2014generative}, we predict a map of probabilistic class scores. 
This has been shown to achieve better results for adapting the appearance of images, for instance by \citet{isola2017image}, whose discriminator architecture we adapt for our purposes. 
Our network, which has about 2.8 M parameters, is described in table~\ref{tab:discriminator}. 
It takes either a transformed image $x_k^{ST}$ or a target domain image $x_k^{T}$ as input and predicts maps of probabilistic class scores $p_T(x^{ST}_k)$ and $p_T(x_k^{T})$, respectively. 
This corresponds to the solid green and yellow paths in figure \ref{fig:overview}. 
Given the receptive field of the network, each value in the probability maps denotes the probability for the corresponding $70 \times 70\,px$ support window in the input image to come from the target domain.
Deviating from \citep{isola2017image}, we replace the batch-normalization layers by a spectral normalization of the weights, as proposed in \citep{spectralnorm}. 
In preliminary experiments we found this to lead to a more realistic appearance of the transformed images.


\begin{table}[!ht]
	\begin{center}
		\begin{tabular}{ l  c  c  c }
			Layer& Layer type & H / W & Depth \\
			\hline
			\hline
			1 & Input layer: MSI + nDSM &  254 & $N$ \\
			2 & Conv(4) stride 2, LReLU &  126 & 64 \\
			3 & SN-Conv(4) stride 2, LReLU &  62 & 128 \\
			4 & SN-Conv(4) stride 2, LReLU &  30 & 256 \\
			5 & SN-Conv(4) stride 1, LReLU &  27 & 512 \\
			6 & SN-Conv(4) stride 1, Sigmoid &  24 & 1 \\
		\end{tabular}
		\caption{\label{tab:discriminator}Discriminator network. SN-Conv($s$): convolution with kernel size $s \times s$ and spectral normalization of the weights. LReLU: leaky ReLU with slope 0.1 (as in \citep{wittich20entmin}). For other abbreviations, cf.~table~\ref{tab:segmentation}.}
	\end{center}
\end{table}


\subsection{Training}
\label{sec:training}

To determine the parameters of all networks, we use a two-stage training strategy which consists of the source training and the subsequent DA. 
In source training, only the parameters $\Theta_C$ of the classification network \textit{C} are determined by conventional supervised training using the labelled source domain dataset $T^S$ (cf.~section \ref{sec:ace}), resulting in the parameter set $\hat{\Theta}_C$.
In the second stage, described in section~\ref{sec:appearance_adaptation}, the actual DA is carried out using $T^S$ and $U^T$. 
In this process, the parameter set $\hat{\Theta}_C$ is used as initialization for $C$, and the parameters of the other networks are randomly initialized according to \citep{he2015delving}.
The second stage could also be carried out starting from a random initialization.
However, in preliminary experiments we observed random initialization to increase the time required for training by a factor of  2.


\subsubsection{Supervised source training}
\label{sec:ace}


During source training, the parameters $\Theta_C$ for the classification network \textit{C} are iteratively updated by stochastic gradient descent, minimizing a supervised loss $\mathcal{L}_{sup}$ that is based on the discrepancy between the predicted labels and the reference. 
Commonly, the cross-entropy loss is used for a multi-class classification problem. 
However, this loss can be suboptimal if the class-distribution of the training dataset is imbalanced, as it is often the case in RS applications. 
In such a case, the cross-entropy loss is dominated by the frequent classes, which is the reason why the prediction quality of under-represented classes may be not satisfactory after training. 
For the reasons discussed in section~\ref{sec:relwork-ID}, we think that existing methods to mitigate this problem are not optimal and, thus, we propose a new approach. 
Similarly to the focal loss  \citep{lin2017focal,yang2019focal}, it also adapts the loss function by a weight that depends on the quality of the prediction. 
However, in contrast to the core idea of the focal loss, the training procedure should not focus on individual pixels whose class labels are difficult predict, but it should focus on $classes$ which are  predicted with low quality. 
Thus, we determine one weight per class (rather than per pixel as in \citep{lin2017focal}) which depends on the current prediction quality of that class. 


After initialization, training starts with one epoch based on the standard softmax-cross-entropy loss. 
In all subsequent training epochs, the training images of a minibatch are classified using the current state of the classifier and the results are compared to the reference to determine the intersection over union $IoU_c$ as a class-specific performance indicator for all \textit{l} classes $c \in \mathbb{L}$: 
\begin{equation}\label{eq:IoU}
IoU_c = \frac{TP_c}{TP_c+FP_c+FN_c}.
\end{equation}
In equation~\ref{eq:IoU}, $TP_c$, $FP_c$ and $FN_c$ are the numbers of true positives, false positives and false negatives, respectively, of all samples assigned to class $c$. 
These IoU scores are used to determine the class-wise weights $w_c$: 
\begin{equation}\label{eq:weights}
w_c = (1 - \Delta IoU_c)^\kappa = [1- (IoU_c - \frac{1}{l}\sum_{h=0}^{l} IoU_h)]^\kappa. 
\end{equation}
In equation~\ref{eq:weights},  $\Delta IoU_c$ is the difference between the class wise $IoU$ score and the mean $IoU$ of all classes, and the hyper-parameter $\kappa$  scales the influence of classes with a lower $IoU$. 
In the next epoch the classifier is trained using a weighted cross-entropy loss in which the loss of each pixel is weighted by $w_c$ according to its reference label. 
For a batch of $B$ training images with height \textit{H} and width \textit{W}, this loss becomes
\begin{equation}\label{eq:Lsup}
\mathcal{L}_{sup}(\Theta_C) = -\frac{1}{N_p} \sum_{b = 1}^{B}\sum_{i = 1}^{H}\sum_{j = 1}^{W}\sum_{c = 1}^{l}  \bar{y}_{b}(i,j,c) \cdot log(r^S_b(i,j,c) ) \cdot w_c,
\end{equation}
where $b$ is the index of an image in the batch, $i, j$ are the indices of a pixel in an image and $c$ is the class index. 
$N_p=H\cdot W \cdot B$ is a short-hand for the number of pixels in a batch. 
The symbol $\bar{y}_b(i,j,c)$ indicates whether pixel $(i, j)$ in the $b^{th}$ label map belongs to class $c$ ($\bar{y}_b(i,j,c) = 1$) or not ($\bar{y}_b(i,j,c) = 0$), whereas $r^S_b(i,j,c)$ denotes the softmax output for class $c$ at  pixel $(i, j)$.
After each new epoch the weights are re-computed according to equations \ref{eq:IoU} and \ref{eq:weights} and used for the subsequent epoch. 
To prevent the classifier from overfitting, we use L2-regularization of the parameters and apply data augmentation in the way described in section~\ref{sec:setup:settings}.


\subsubsection{Embedded appearance adaptation} 
\label{sec:appearance_adaptation}


In the adaptation phase, the parameters of all networks are determined using variants of stochastic gradient descent.
To simplify the notation we define a combined set $\Theta_{A,C} = \{\Theta_A,\Theta_C\}$ consisting of the parameters of the adaptation network \textit{A} and the classification network \textit{C}. 
As we want to determine these parameters by joint training of the two networks, according to the principles of adversarial training, in each iteration the gradient of a joint loss function $\mathcal{L}_{A,C}$ with respect to $\Theta_{A,C}$ is computed first.
This corresponds to the paths visualized by the red, the blue and the two green arrows in figure \ref{fig:overview}, based on a batch of \textit{B} images from $\mathcal{D}^{S}$ and the corresponding labels.
The resulting gradient is used to update the parameter set $\Theta_{A,C}$.
After that, within the same training iteration, the gradient of a discriminator loss $\mathcal{L}_{D}$ with respect to the parameters $\Theta_D$ of the discriminator is computed and used to update $\Theta_D$. 
This requires an additional set  of \textit{B} unlabelled images from $\mathcal{D}^{T}$, which are processed jointly with the \textit{B} transformed images from $\mathcal{D}^{S}$ that contributed to the update of $\Theta_{A,C}$, and it corresponds to the paths visualized by the yellow and green arrows in figure \ref{fig:overview}. 
We follow the common practice to alternate between updating $\Theta_{A,C}$ and $\Theta_D$ \citep{goodfellow2014generative}.
In the following, the two steps are described in detail.


\paragraph{{\bf Joint update of $A$ and $C$}}
The joint loss $\mathcal{L}_{A,C}$ used to update \textit{A} and \textit{C} consists of three components:
\begin{equation}\label{eq:Lcla}
\mathcal{L}_{A,C}(\Theta_{A,C},\Theta_D) =  \omega_{T} \cdot \mathcal{L}^{ST}_{sup}(\Theta_{A,C}) + \mathcal{L}_{sup}(\Theta_C) + \omega_G \cdot \mathcal{L}_{advA}(\Theta_{A},\Theta_{D}), 
\end{equation}
where $\omega_{T}$ and $\omega_{G}$ are weighting factors to control the relative influence of the corresponding  loss terms. 


The first term $\mathcal{L}^{ST}_{sup}$ is related to the main goal of the DA, namely to achieve a good classification performance on images from the target domain.
It is formulated as a supervised classification loss for transformed images similarly to equation~\ref{eq:Lsup}: 
\begin{equation}\label{eq:LsupT}
\mathcal{L}^{ST}_{sup}(\Theta_{A,C}) = -\frac{ 1}{N_p} \sum_{b = 1}^{B}\sum_{i = 1}^{H}\sum_{j = 1}^{W}\sum_{c = 1}^{l}  \bar{y}_{b}(i,j,c) \cdot log(r^{ST}_b(i,j,c) ) \cdot w_c, 
\end{equation}
where  $r^{ST}_b(i,j,c)$ denotes the predicted probabilistic class score for the pixel $i,j$ in the $b^{th}$ transformed input image $x^{ST}$ and the remaining symbols are those already defined in the context of equation~\ref{eq:Lsup}.


The second term in equation~\ref{eq:Lcla}, $\mathcal{L}_{sup}(\Theta_C)$ is the supervised loss for images from the source domain and is computed according to equation \ref{eq:Lsup}. 
This is important to achieve semantic consistency. 
If \textit{C} were solely trained using the transformed images, it would be possible for the appearance adaptation network to produce semantically inconsistent results, e.g. images in which transformed trees look like target domain buildings and vice versa. 
In this case, $C$ could still learn to predict the label maps correctly,  but it would no longer perform well for real target images. 
Considering $\mathcal{L}_{sup}(\Theta_C)$ for source images will constrain the target classifier so that does not deviate too much from the classifier of the source domain, so that this loss acts as a regularization term. 


The last term in equation \ref{eq:LsupT} is the adversarial loss $\mathcal{L}_{advA}$ for $A$ and realizes the component of adversarial training that influences $A$. 
The appearance adaptation module \textit{A} should learn to transform images from $\mathcal{D}^S$ such that they look like images from $\mathcal{D}^T$ by maximising the probabilities predicted by \textit{D} for the transformed images $x^{ST}$  to be from $\mathcal{D}^T$, which results in the following loss: 
\begin{equation}\label{eq:Lgen}
\mathcal{L}_{advA}(\Theta_A,\Theta_D) = -\frac{1}{N_p'} \sum_{b = 1}^{B}\sum_{i = 1}^{H'}\sum_{j = 1}^{W'}  log(p_T(x^{ST}_b)(i,j)),
\end{equation}
where $p_T(x^{ST}_b)(i,j)$ corresponds to the prediction of \textit{D} at position ($i,j$), i.e. the predicted probability of the corresponding support window in the $b^{th}$ transformed input image $x^{ST}_b$ presented to \textit{D} to be an image from the target domain, which should be large such that $A$ can learn to fool the discriminator (cf.~section~\ref{sec:methodology:network:discriminator}).  
$H'$ and $W'$ are the height and width of the discriminator output, respectively, and $N_p' = B \cdot H'\cdot W'$ is the number of discriminator pixels in the batch with $B$ images. 


As stated above, each training iteration starts with a step aiming at minimizing the joint loss $\mathcal{L}_{A,C}$ with respect to $\Theta_A$ and $\Theta_C$. 
While  $\mathcal{L}_{A,C}$  depends on $\Theta_D$ via the adversarial term $\mathcal{L}_{advA}$, these parameters are not updated in this context. 
Minimizing $\mathcal{L}_{A,C}$ will adapt the adaptation network $A$ such that fools the discriminator $D$, i.e. such that its output cannot be discriminated from a real target domain image, and at the same time it will update the classifier $C$ such that it performs well for target domain images. 
The supervised loss $\mathcal{L}_{sup}$ acts as a regularizer for $\Theta_C$. 


In one training iteration, a source domain image will contribute to the loss $\mathcal{L}_{A,C}$ twice, namely via $\mathcal{L}^{ST}_{sup}$ (equation~\ref{eq:LsupT}) and $\mathcal{L}_{sup}$ (equation \ref{eq:Lsup}). 
This has to be considered in the batch normalization layers of $C$. 
We only use the transformed images to compute the running averages of these layers, because we assume them to be more closely related to the statistics of $\mathcal{D}^T$ than the values for the source domain images if the appearance adaptation is successful.

\paragraph{{\bf Update of $D$}} 
The second step of adversarial training is related to the update of the discriminator network $D$, which is based on a loss function $\mathcal{L}_{D}$ consisting of two terms, the second one being weighted by $\rho$: 
\begin{equation}\label{eq:Ldiss}
\mathcal{L}_{D}(\Theta_A,\Theta_D) = \mathcal{L}_{advD}(\Theta_A,\Theta_D) + \rho \cdot \mathcal{L}_{reg}(\Theta_A,\Theta_D). 
\end{equation} 


The first term is typical for adversarial training and is supposed to train the discriminator network \textit{D} to differentiate real images $x^T$ from $\mathcal{D}^T$ and transformed source images $x^{ST}$: 
\begin{equation}\label{eq:Ldis}
\mathcal{L}_{advD}(\Theta_A,\Theta_D) = -\frac{1}{N_p'} \sum_{b = 1}^{B}\sum_{i = 1}^{W'}\sum_{j = 1}^{H'} log(p_T(x^T_b)(i,j)) + log(1-p_T(x^{ST}_b)(i,j)), 
\end{equation}
where $p_T(x^T_b){(i,j)}$ is the probability for the image patch corresponding to the discriminator pixel ($i, j$)  for the $b^{th}$ image in a batch of images from $\mathcal{D}^T$ to be obtained from a target domain image and the remaining symbols are identical to those defined in the context of equation~\ref{eq:Lgen}.


The second term in equation \ref{eq:Ldiss} corresponds to the proposed new regularization for \textit{D} that is motivated by the following line of thought, supported by observations in preliminary experiments. 
Whenever the feature distributions between $\mathcal{D}^S$ and $\mathcal{D}^T$ are very different, the discriminator network can easily distinguish the domains by simply focussing on frequent features. 
For instance, if $\mathcal{D}^T$ has a higher frequency of pixels corresponding to vegetation than $\mathcal{D}^S$, the discriminator will quickly learn to predict the probability for an image to be drawn from $\mathcal{D}^T$ based on the number of pixels that are representative for vegetation. 
In consequence, \textit{D} will predict such regions to come from $\mathcal{D}^T$ with a high probability, and this will have a higher impact on the overall decision for an image than other areas which are more difficult to differentiate and where the predicted probabilities might be close to $50\,\%$.
We note that such a situation results in a high variance in the predicted maps $p_T(\cdot)$.
As \textit{A} tries to fool the discriminator, it will learn to mimic such features, e.g. to predict vegetation areas in the example just mentioned.
However, as the respective areas may correspond to other classes, this would results in hallucinated structures, which means that semantic consistency will not be achieved. 
To prevent the discriminator from learning such solutions, we propose to constrain the variance of the output of the discriminator. 
In this way, \textit{D} also has to learn to also differentiate the more difficult areas and, thus, to learn non-trivial differences between the domains. 
To do so, the batch-wise standard deviation of the values of the discriminator output $p_T(\cdot)$ for the target images and the transformed source images of a batch is penalized, which results in the regularization loss
\begin{equation}\label{eq:Lreg}
\mathcal{L}_{reg}(\Theta_A,\Theta_D) = \sum_{Q\in{S,ST}}\sqrt{\frac{1}{N_p'-1}\sum_{b=1}^{B}\sum_{i=1}^{W'}\sum_{j=1}^{H'}(     p_T(x^Q_b){(i,j)}-\bar{p}_T^Q)^2},
\end{equation} 
where $\bar{p}_T^Q$ denotes the average value in $p_T(x^Q)$ for the image set $Q$ in the batch. 
The remaining symbols are as described in the context of equation~\ref{eq:Ldis}.


In order to update the parameters of $D$, the gradients of $\mathcal{L}_{advD}(\Theta_A,\Theta_D)$ with respect to $\Theta_D$ are used, which will train $D$ such that it can discriminate well between real target images and transformed source images, thus making the task of the adaptation network $A$ more difficult. 
Note that in this process, the transformed images are not directly presented  to the discriminator, but instead a small horizontal and vertical shift as well as a radiometric transformation as  described in section \ref{sec:setup:settings}   are applied. 
The shift is drawn from a uniform distribution with the lower limit of zero and the upper limit of 4 pixels, which corresponds to the filter size of the first convolution in \textit{D}. 
In preliminary experiments, we found this to reduce artefacts in the form of high-frequency repeating patterns in the transformed images. 


\subsection{Entropy based parameter selection} 
\label{sec:eps}


Many DA methods rely on iterative processes that are repeated for a fixed number of epochs, using the parameter set after the very last iteration for inference \citep{tasar2020semI2I, tasar2020colormapgan, benjdira2019unsupervised}. 
For the reasons given in section~\ref{sec:relwork-PS} we propose another strategy, namely to select  a parameter set according to an optimality criterion derived from the data. 
For lack of labelled data in  $\mathcal{D}^T$, this criterion cannot be based on the validation error in that domain.
Instead, we use the average entropy of the predicted class scores for images $x^T$ from $\mathcal{D}^T$ as an approximate measure for the validation performance. 
The entropy of the class scores can be interpreted as a measure of uncertainty of the predictions \citep{wittich20entmin}. 
Consequently, we expect good parameter values to lead to a low uncertainty of the predictions and, thus, to a low entropy. 
After each epoch in the adaptation, the classification network \textit{C} is used to predict the probabilistic class scores $r^{T} = (r_1,..,r_c,..,r_l)$ for every pixel of all images from $U^T$. 
For such a pixel, the entropy $E$ is computed according to
\begin{equation}\label{eq:ent}
E = -\frac{1}{log(l)}\sum_{c=1}^{l}r_c\cdot log(r_c), 
\end{equation} 
and the average entropy $\bar{E}$ is determined from all the pixel-wise values.  
The parameter set having the lowest value of $\bar{E}$ is selected after running the adaptation for a fixed number of epochs. 
To save computation time, the mean entropy is not computed for the first couple of epochs, because \textit{A} and \textit{D} are not expected to deliver meaningful results in the beginning of the adaptation phase.


\subsection{Resolution adaptation}
\label{sec:resolutiontransfer}


Methods for appearance adaptation have problems if the ground sampling distance (GSD) of the source domain ($GSD^S$) is different from the one of the target domain ($GSD^T$) \citep{liu19curve,benjdira2019unsupervised}. 
To overcome this problem we pre-scale the images of one domain using the information about the GSD, which is usually available in RS applications. 
If $GSD^T > GSD^S$ the data from $\mathcal{D}^S$ are downsampled to the resolution of $\mathcal{D}^T$. 
This includes downsampling of both, the image data (using bilinear interpolation) and the reference (using nearest neighbour interpolation). 
Then, source training and DA are performed using the resampled data from $\mathcal{D}^S$ and the original data from $\mathcal{D}^T$. 
On the other hand, if $GSD^T < GSD^S$, we downsample the image data from $\mathcal{D}^T$ to $GSD^S$ and perform source-training and domain adaptation in the resolution of $\mathcal{D}^S$. 
In order to predict label maps for $\mathcal{D}^T$ at the original resolution, the predicted probabilities $r_k^{ST}$ are upsampled to  $GSD^T$ using bilinear interpolation. 
The pixel-wise class predictions are then obtained by selecting the class with the highest probability for each pixel in the upsampled probability maps.

\section{Experiments}
\label{sec:experiments}


\subsection{Test Datasets}
\label{sec:datasets}


For the evaluation of the proposed method seven datasets were used, each showing a different German city and each being treated as a single domain. 
The first group of datasets consists of MSI, height data and label maps for the cities of Schleswig (\textit{S}), Hameln (\textit{Hm}), Buxtehude (\textit{B}), Hannover (\textit{H}) and Nienburg (\textit{N}) with a GSD of 20 cm  \citep{wittich20entmin}. 
All of these datasets include $\mathtt{RGBI}$ images (red, green, blue, near infrared) and a normalized digital surface models (nDSM) which contains the height above ground for each pixel. 
The blue channel was not used in the experiments, because it was not available for all datasets.
The reference was generated by manual labelling according to the class structure shown in table~\ref{tab:datasets}. 
In \citep{wittich20entmin}, each dataset was split into subsets for training and testing, respectively. 
We denote the training subset of a city $D$ by $\dot{D}$ and the testing subset as $\tilde{D}$.


\begin{table}[ht]
	\begin{center}
		\begin{tabular}{ c l c | c  c  c c  c c c }
			&\multicolumn{2}{l|}{\textbf{City}} & $P_{20}$ & $V_{20}$ & \textit{S} & \textit{Hm} & \textit{B} & \textit{H} & \textit{N} \\
			\hline
			&\multicolumn{2}{l|}{Size in M pixel} & 85.5 & 34& 26 & 37& 100& 100& 100\\
			&\multicolumn{2}{l|}{Capturing season} & A & S& S & A & S & S & S \\
			\hline
			\multirow{6}{*}{\rotatebox[origin=c]{90}{\parbox[c]{2.7cm}{\centering Class distr. [\%]}}}
			&Sealed Ground   &(SG)   & 29.6 & 27.8 & 14.1 & 18.8 & 22.1 & 33.6& 22.8 \\
			&Building        &(BU)   & 25.7 & 26.0 & 14.7 & 19.1 & 19.7 & 36.7& 18.4 \\
			&Low Vegetation  &(LV)   & 22.6 & 21.3 & 38.9 & 36.2 & 36.9 &  7.5& 40.3 \\
			&High Vegetation &(HV)   & 15.5 & 22.9 & 31.5 & 24.5 & 20.3 & 20.6& 17.8 \\
			&Vehicle         &(VH)   &  1.8 &  1.2 &  0.8 &  1.3 &  1.0 &  1.6&  0.7 \\
			&Clutter         &(CL)   &  4.8 &  0.8 &   -  &   -  &   -  &  -  &   -  \\
		\end{tabular}
		\caption{\label{tab:datasets} Dataset overview. Capturing season is either autumn (A) or summer (S). Class distr.: percentage of pixels assigned to the corresponding class in every city. }
	\end{center}
\end{table}


We also used the Potsdam (\textit{P}) and Vaihingen (\textit{V}) datasets of the ISPRS labelling benchmark \citep{wegner2017}. 
They consist of ortophotos, nDSMs and label maps with 6 classes as shown in table \ref{tab:datasets}. 
\textit{P} consists of $\mathtt{RGBI}$ images  captured with a GSD of 5 cm, whereas the imagery from \textit{V} has a GSD of 9 cm and does not include a blue channel. 
Both datasets were split into training and testing areas by the benchmark organizers; 
we denote the training areas by $\dot{P}$ and $\dot{V}$ and the test areas by $\tilde{P}$ and $\tilde{V}$, respectively. 
In the experiments conducted to compare our method to \citep{liu19curve}, we further use a subset $\bar{V}$ which includes areas from both, $\dot{V}$ and $\tilde{V}$; 
cf.~\citep{liu19curve} for details.
In some experiments, we used resampled versions of the datasets $P$ and $V$, which is indicated by a subscript showing the GSD, e.g. $V_{20}$ refers to data from $V$ resampled to a GSD of 20 cm.
We use bilinear interpolation for resampling the image and height data and nearest neighbour interpolation for the label maps.
In order to align the class structure of $P$ and $V$ to the one of the other cities, we follow the approach of \citet{liu19curve} and ignore the class \textit{Clutter}.  
The corresponding regions in the reference do not contribute to the training loss, and at test time they do not contribute to the evaluation procedure. 
Whenever \textit{Clutter} is ignored, we denote the datasets by $P'$ and $V'$, respectively.


Table~\ref{tab:datasets} shows the class distributions as well as the overall size of each dataset, along with the capturing season as a possible reason for a domain gap. 
In all datasets the class \textit{Vehicle} is strongly under-represented. 
Also, the class distributions are very different between the datasets.
For example, $S$ has a larger amount  of \textit{High Vegetation} ($31.5\,\%$) than $P$ ($15.5\,\%$), and $H$ has a much smaller amount of \textit{Low Vegetation}  ($7.5\,\%$) than $B$ ($36.9\,\%$). 
The Jensen-Shannon-Divergence ($D_{JS}$) \citep{lin1991divergence} indicates that the class distributions of $S$ and $H$ are the least similar ones (($D_{JS} = 0.33$), while the distributions of $N$ and $B$ are most similar ($D_{JS} = 0.04$). 
Each dataset was pre-processed so that each channel has a mean of zero and a standard-deviation of one. 
In case of the nDSMs, the heights were divided by a constant value of 30 m (instead of the standard deviation) to preserve the relative metric height information. 


\subsection{Experimental settings, evaluation protocol and test setup}
\label{sec:setup}


\subsubsection{Experimental settings}
\label{sec:setup:settings}


In all experiments, the supervised source training of the classification network is conducted using stochastic minibatch gradient descent with a learning rate of $0.01$ and momentum of $0.9$. 
We use a L2-regularisation of $\Theta_C$, implemented as weight decay with a weight of $1e^{-5}$. 
These values were adopted from  \citep{xception}, the only difference being we do not decrease the learning rate over time but instead start with a lower learning rate, which is a common strategy when starting from a pre-trained network. 
Further hyper-parameters were  tuned empirically using the domain $N$ by training multiple networks with different sets of hyper-parameters and selecting the parameter set achieving the highest mean F1-score on  $\tilde{N}$. 
This domain was chosen because it was also used for tuning in \citep{wittich20entmin}, to which we compare our new method in one of the experiments.
As a result, the parameter of the adaptive loss (equation~\ref{eq:weights}) is set to $\kappa = 4$ and the number of epochs of source training to 50, where each epoch consists of 2,500 iterations. 
The batch-size was set to $B=4$.


The hyper-parameters for DA were tuned by performing DA from \textit{S} to \textit{Hm} with different parameter values and choosing the ones achieving the highest mean F1-score on \textit{Hm} after the adaptation. 
This pair of domains was chosen because it is very challenging, \textit{S} having been captured in summer and \textit{Hm} in autumn. 
The resulting weights $\omega_T$ and $\omega_G$ in equation~\ref{eq:Lcla} were set to $2$ and the weight of the regularization term in equation \ref{eq:Ldiss} was set to $\rho=4$. 
The batch-size was again set to $B=4$.
Following \citep{isola2017image,soto2020domain}, the appearance adaptation network \textit{A} and the discriminator \textit{D} are both optimized using the ADAM optimizer \citep{kingma2014adam} using $\beta_1 = 0.9$ and $\beta_2 = 0.99$ and a learning rate of $1e^{-4}$. 
The adaptation is run for 50 epochs. 
After epoch 25 we start with evaluating the entropy based selection criterion (cf.~section~\ref{sec:eps}), so that the parameter set achieving the lowest average entropy of the class scores after epoch 25 defines the parameter set to be used for classification.


We apply data augmentation to images from both domains in the source training and DA procedures, following the strategy presented in \citep{wittich20entmin}. 
We crop randomly rotated patches out of all available training images to build the training batches online. 
While the rotated patches are generated using bilinear interpolation, the corresponding label maps are generated using nearest-neighbour interpolation. 
Each channel of the cropped and rotated patches is  multiplied by a random value drawn from a normal distribution $\mathcal{N}(\mu=1, \sigma)$, and a random value drawn from $\mathcal{N}(\mu=0,\sigma)$ is added. 
We used $\sigma=0.1$ in both cases; this value was determined in the parameter tuning process. 

To classify an image that is larger than the input size of \textit{C}, we use an inference protocol that aims at increasing the quality of the predictions \citep{wittich20entmin}.
The image is processed by \textit{C} in a sliding window fashion with an horizontal and vertical overlap of $128\,px$. 
To further increase the redundancy, each window is additionally flipped in both, horizontal  and vertical directions,  and it is also rotated by 180$^o$; these flipped and rotated versions of the window are classified, too. 
In the end, all class scores that correspond to the same pixel are averaged and the class achieving the highest score is selected as the classification result.


\subsubsection{Evaluation protocol}
\label{sec:setup:eval}

Based on the predicted label maps, a confusion matrix is determined by comparing the predictions to the reference label maps. From the confusion matrix, the number of true positive ($TP_c$), false positive ($FP_c$) and false negative ($FN_c$) predictions are derived for each class $c \in \mathbb{L}$ . 
For each class, we compute the intersection over union $IoU_c$ (cf.~equation~\ref{eq:IoU}) and the  F1-score $F_{1,c}$: 
\begin{equation}\label{eq:F1}
 F_{1,c}= \frac{TP_c}{TP_c + 0.5 \cdot (FP_c + FN_c)}. 
\end{equation}
As global metrics, the overall accuracy $OA$, i.e. the percentage of correct class assignments, the  mean intersection over union $\overline{IoU}$ and the mean F1-score $\bar{F_1}$ are reported, where the means are taken over  all classes. 
In addition, we consider the positive transfer rate $PTR(M)$, where $M$ can be any of the above performance metrics.
We present this metric as $PTR(M) = n_{pt}/n_{exp}$ where $n_{exp}$ is the number of different adaptation scenarios in an experiment and $n_{pt}$ the number of positive transfers w.r.t. the performance metric $M$.


\subsubsection{Test setup}
\label{sec:setup:testsetup}

Using the parameter settings and training methods  described  in section~\ref{sec:setup:settings}, different experiments were conducted that are presented in the subsequent sections. 
In all these experiments, a domain corresponds to the data of one of the cities described in section~\ref{sec:datasets}. 
The first set of experiments solely evaluates source training and mainly provides a baseline for the other experiments. 
On the one hand, evaluating the classification accuracy that can be achieved when the training and test data belong to the same domain is an indication for what could be expected in the optimal case; on the other hand, the results achieved when applying a classifier trained on source domain data to the target domain without DA corresponds to the worst-case scenario and indicates the domain gap for each combination of cities serving as source and target domains, respectively. 
These experiments are reported in  section~\ref{sec:adaptation-1}. 
Section~\ref{sec:adaptation-2} is dedicated to the evaluation of domain adaptation. 
As seven domains are available, there are 42 possible pairs of source and target domains, and for every combination we report the evaluation metrics achieved in the target domain after the adaptation. 
After that, several ablation studies are reported in  section~\ref{sec:ablation}, in which we want to assess the influence of some components of our method on the classification results. 
Finally, in section~\ref{sec:comparative}, our method is compared to existing DA approaches from the field of RS. 
Details on the experimental protocols that deviate from those described in section~\ref{sec:setup:settings} and the selection of the datasets for specific experiments are discussed in the respective sections. 


\subsection{Evaluation of source training}
\label{sec:adaptation-1} 


In order to evaluate the source training, we used all datasets described in section~\ref{sec:datasets} at a \textit{GSD} of $20\,cm$. 
In case of $P$ and $V$, the results are based on the datasets $P'_{20}$ and $V'_{20}$, respectively, i.e. we excluded the class \textit{Clutter}. 


\subsubsection{Training and testing on the same domain}
\label{sec:adaptation-1-same} 


In each of the experiments reported in this section, we trained a classifier using training data from one city and applied it to test data from the same city, using the definition of the (non-overlapping) training and test sets described in \citep{wittich20entmin}. 
DA was not applied. 
Thus, table \ref{tab:splitted_eval} presents the performance of classifiers trained on the training subset $\dot{O}$ and evaluated on the testing subset $\tilde{O}$ for each city $O \in { P'_{20}, V'_{20},S,Hm,B,H,N}$.


\setlength{\tabcolsep}{0.74em}
\begin{table}[ht]
	\begin{center}
		\small
		\begin{tabular}{ c | c | c | c | c | c | c | c | c  }
			
			&$\tilde{P}'_{20}$  & $\tilde{V}'_{20}$  & $\tilde{S}$  & $\tilde{H}m$  & $\tilde{B}$  & $\tilde{H}$  & $\tilde{N}$ & Avg.  \\
			\hline
			\hline
			 $\bar{F_1}$ [\%] &$88.3$ & $83.8$ & $83.6$ & $86.9$ & $86.9$ & $80.6$ & $86.2$ & $85.2$ \\
			 $OA$ [\%] & $88.9$ &  $86.8$ & $88.4$ & $89.8$ & $88.3$ & $86.4$ & $89.1$ & $88.2$ \\
		\end{tabular}
		\caption{\label{tab:splitted_eval} $\bar{F_1}$ scores and overall accuracies obtained on the test sets $\tilde{O}$ for the domains $O \in { P'_{20}, V'_{20},S,Hm,B,H,N}$ after training on the training subsets of the same domain.}
	\end{center}
\end{table}

As training and test data are taken from the same domain, we expect the results in table~\ref{tab:splitted_eval} to represent the optimum that can be achieved by our classification network $C$.
They are in line with the current state of the art in pixel-wise classification. 
Whereas the highest OA achieved for $V$ and $P$ according to the score board of the ISPRS benchmark \citep{wegner2017} is about 91\%, the benchmark protocol excludes pixels near object boundaries from the evaluation, where wrong classification results are more likely to occur than in other areas. 


\subsubsection{Training and testing on different domains}
\label{sec:adaptation-1-different} 


For the experiments described in this section, the source classifier, trained using the entire dataset from one city ($\mathcal{D}^S$), was applied to all data from another city ($\mathcal{D}^T$). 
This was performed for all possible combinations of source and target domains. 
For all of these 42 DA scenarios, the results show how a classifier performs if it is applied to another domain, and the effectiveness of DA can be assessed by comparing the performance in $\mathcal{D}^T$ after DA to the one reported here.
Note that we did not use the split into training and test samples that was used in section~\ref{sec:adaptation-1-same} because we wanted the evaluation of source training and of DA in the next section to be based on a dataset that was as large as possible. 
Table \ref{tab:pre-training} presents the $\bar{F_1}$ values for this cross-domain evaluation. 
The last column and the last row show the average scores for every source and target domain, respectively.
In the bottom-right corner of the table the average score over all scenarios is presented. 
The result for the scenario $\textit{S}\rightarrow \textit{Hm}$ was excluded in the computation of all average scores because this setting was used for tuning the DA method. 
We do not report other qualtity metrics to save space; they behave very similarly to $\bar{F_1}$. 


The $\bar{F_1}$  values in table \ref{tab:pre-training} vary quite strongly between the scenarios. 
The average performance is worst when applying the models to $H$, which is probably due to the fact that the centre of Hannover is much more densely populated than the other cities. 
A second possible reason is the bad quality of the height information in $H$, where the nDSM has a patchy appearance and height changes are badly aligned with the objects; cf. example for $x^S$ in figure \ref{fig:overview}.
Nevertheless, the model trained on $H$ performs rather good in the other domains; only the models trained on $Hm$ and $N$ perform better on average. 
Again this could be due to the bad quality of the height information which makes the classifier focus on the radiometric information.
A strong influence of seasonal differences between $\mathcal{D}^S$ and $\mathcal{D}^T$ cannot be observed. 
For example, the results for the setting $P'_{20}\rightarrow \textit{Hm}$ are rather bad although the data for both domains were captured in autumn. 
Contrarily, the results for the opposite scenario $\textit{Hm}\rightarrow P'_{20}$ are comparatively good, which also indicates that the domain gap is not symmetrical.


\setlength{\tabcolsep}{0.74em}
\begin{table}[ht]
	\begin{center}
		\small
		\begin{tabular}{ c || c | c | c | c | c | c | c || l }

			\scriptsize \backslashbox{$\mathcal{D}^S$}{$\mathcal{D}^T$} & $P'_{20}$  & $V'_{20}$  & \textit{S}  & \textit{Hm} & \textit{B}  & \textit{H}  & \textit{N} & Avg. \\
			\hline
			\hline
			$P'_{20}$  &    -   & $77.0$ & $65.7$ & $68.9$ & $76.6$ & $59.3$ & $75.6$ & $70.5$\\
			\hline
			$V'_{20}$  & $74.9$ &    -   & $72.8$ & $69.4$ & $80.8$ & $52.1$ & $78.6$ & $71.4$\\
			\hline
			\textit{S}  & $70.6$ & $76.6$ &    -   &($78.5$)& $79.4$ & $57.6$ & $74.4$ & $71.7$\\
			\hline
			\textit{Hm} & $78.0$ & $79.0$ & $81.7$ &    -   & $82.8$ & $73.8$ & $79.1$ & $79.2$\\
			\hline
			\textit{B}  & $70.3$ & $81.7$ & $81.4$ & $72.9$ &    -   & $58.9$ & $78.4$ & $73.9$\\
			\hline
			\textit{H}  & $76.1$ & $76.8$ & $67.1$ & $76.5$ & $78.9$ &    -   & $74.0$ & $74.9$\\
			\hline
			\textit{N}  & $78.0$ & $81.5$ & $78.6$ & $76.9$ & $83.5$ & $67.4$ &    -   & $77.7$\\
			\hline
			\hline
			Avg.        & $74.7$ & $78.8$ & $74.6$ & $72.9$ & $80.3$ & $61.5$ & $76.7$ & $74.2$ \\

		\end{tabular}
		\caption{\label{tab:pre-training}Mean F1 scores $\bar{F_1}$ obtained on $\mathcal{D}^T$ after source training on $\mathcal{D}^S$ (before DA). Avg.: average scores per row / column, respectively, excluding the scores for the scenario with $S$ as source and $Hm$ as  target domains.}
	\end{center}
\end{table}


Representing results before DA, the metrics in table~\ref{tab:pre-training} correspond to the worst-case scenario that can occur when a classifier is transferred to another domain without adaptation. 
A direct comparison of these results to the best-case scenario described in section~\ref{sec:adaptation-1-different} and especially in table \ref{tab:splitted_eval} is not possible because the numbers are based on a different test set (test set  $\tilde{O}$ of a domain $O$ in table \ref{tab:splitted_eval}, all data from domain $O$  in table~\ref{tab:pre-training}). 
Nevertheless, we assume the results in table  \ref{tab:splitted_eval} to be a good approximation of the missing values on the main diagonal in table \ref{tab:pre-training}.
Comparing the average $\bar{F_1}$  values of the intra-domain settings in table \ref{tab:splitted_eval} ($85.9\,\%$) to the ones of the cross-domain settings in table \ref{tab:pre-training} ($74.2\,\%$), we conclude that there is a considerable performance drop of $11.7\,\%$ on average, which we attribute to the domain gap between the datasets.


\subsection{Evaluation of DA}
\label{sec:eval_of_da}\label{sec:adaptation-2} 


Using the  parameters after source training (section~\ref{sec:adaptation-1-different}) as initialization for $C$, the proposed DA method is used to adapt each model to all domains but the source domain. 
The adapted models are then evaluated in $\mathcal{D}^T$. 
Table \ref{tab:adaptation} shows the $\bar{F_1}$  values after DA. 
Analogously to table \ref{tab:pre-training}, the last column and the last row show the average metrics by source and target domain, respectively, and the value in the bottom-right corner shows the average score over all scenarios. 
Again, the result for \textit{S}$\rightarrow$\textit{Hm} was excluded in the computation of the averages. 
For an easier comparison, the improvements of the averaged metrics after DA over the scores after source-training in table \ref{tab:pre-training} are listed in parenthesis.


\setlength{\tabcolsep}{0.3em}
\begin{table}[ht]
	\small
	\begin{center}
		\begin{tabular}{ c || c | c | c | c | c | c | c || c }

			\scriptsize \backslashbox{$\mathcal{D}^S$}{$\mathcal{D}^T$} & $P'_{20}$   & $V'_{20}$  & \textit{S}  & \textit{Hm} & \textit{B}  & \textit{H}  & \textit{N} & Avg. \\
			\hline
			\hline
			$P'_{20}$  &    -   & $79.8$ & $72.3$ & $78.9$ & $79.5$ & $69.3$ & $78.6$ & $76.4 \; (+5.9)$\\
			\hline
			$V'_{20}$  & $78.0$ &    -   & $79.9$ & $77.3$ & $82.6$ & $68.7$ & $81.1$ & $77.9 \; (+6.5)$\\
			\hline
			\textit{S} & $72.2$ & $78.6$ &    -   &($82.4$)& $81.1$ & $66.0$ & $77.0$ & $75.0 \; (+3.3)$\\
			\hline
			\textit{Hm}& $78.8$ & $81.3$ & $82.9$ &    -   & $83.7$ & $77.7$ & $80.1$ & $80.8 \; (+1.7)$\\
			\hline
			\textit{B} & $75.7$ & $84.0$ & $84.3$ & $76.3$ &    -   & $72.2$ & $82.2$ & $79.1 \; (+4.2)$\\
			\hline
			\textit{H} & $76.4$ & $79.7$ & $76.3$ & $82.0$ & $80.8$ &    -   & $77.8$ & $78.8 \; (+3.9)$\\
			\hline
			\textit{N} & $79.1$ & $83.2$ & $82.5$ & $80.5$ & $85.9$ & $73.2$ &    -   & $80.7 \; (+3.0)$\\
			\hline
			\hline
			Avg.       & $76.7$ & $81.1$ & $79.7$ & $79.0$ & $82.3$ & $71.2$ & $79.5$ & $78.5$\\  
			           &$(+2.0)$&$(+2.3)$&$(+5.1)$&$(+6.1)$&$(+2.0)$&$(+9.7)$&$(+2.8)$&$(+4.3)$
			
		\end{tabular}
		\caption{\label{tab:adaptation} $\bar{F_1}$ obtained on $\mathcal{D}^T$ after adapting from $\mathcal{D}^S$. The values in parenthesis show the relative improvements compared to source training in table \ref{tab:pre-training}. 
		}
	\end{center}
\end{table}


In all scenarios, DA results in a positive transfer, i.e. the mean F1-score $\bar{F_1}$  is higher after DA than before DA. 
In all cases, the \textit{OA} and the mean \textit{IoU} are improved, too, and behave equivalently in a qualitative sense, but as in section \ref{sec:adaptation-1}, we only report  $\bar{F_1}$ to save space. 
The improvement of  $\bar{F_1}$ due to DA ranges from a minimum of $0.3\,\%$ for the setting \textit{H}$\rightarrow P'_{20}$ to a maximum of $16.6\,\%$ for $V'_{20}\rightarrow$\textit{H}. 
In general, the improvement is higher when the initial performance before DA was lower. 
This seems understandable because improving better models is more difficult. 
When comparing the average improvements by domains it can be seen that the adaptation from and to $H$ resulted in the largest average improvement. 
The probable reason for this is a relatively large visual difference of this domain from all he others due to the different density of urban development as well as the special appearance of the height information in $H$ (cf.~section~\ref{sec:adaptation-1-different}).
Nevertheless, even after adaptation the average performance on \textit{H} is still considerably worse with respect to $\bar{F_1}$ compared to the other domains. 
The models that were originally trained on \textit{Hm} have on average the best performance before and after adapting them to the other domains, but DA achieves the smallest improvement. 
We conclude that our approach results in a stable improvement and that it is higher if a model  has a poorer performance in $\mathcal{D}^T$ before DA.


In the scenarios $S\rightarrow H$ and $H\rightarrow S$, involving the domains with the most dissimilar class distributions with $D_{JS}=0.33$ (cf.~section~\ref{sec:datasets}), an improvement of about $9\,\%$ could be achieved by DA, which is about twice the average improvement over all scenarios. 
However, in both cases the resulting performance after DA is lower than the one achieved in the same target domains when adapting from almost all other domains (the exception being  $P'_{20}\rightarrow S$, which has an even lower $\bar{F_1}$ value than $H\rightarrow S$). 
$B$ and $N$ are domains with most similar class distributions. 
Of all scenarios involving $N$ as target domain, the adaptation based on $B$ as source domain performs best. 
Vice versa, using  $N$ as source domain results in the best performance of all scenarios involving $B$ as target domain.
We take these results as an indication that a large difference in the class distributions does indeed lead to a worse performance in the target domain after DA. 
However, as DA improves the result in the respective scenarios considerably, we think that this is mainly due to a poor initial performance of the classifiers in the target domains before DA. 
We also believe that different class distributions are likely to go along with a different appearance of objects. 
For example, buildings in densely developed urban areas look different from and cover a larger area than those in suburban areas, which can lead to a bad initial performance in the target domain.


To further summarize the DA performance, the average performance metrics for the cross-domain scenarios before and after DA are presented in table \ref{tab:adaptation-global}, again not considering  the results for $\textit{S}\rightarrow \textit{Hm}$.
Again we can compare the average $\bar{F_1}$ values achieved  after DA in table \ref{tab:pre-training}, which is $78.5\,\%$, to the one achieved in the intra-domain setting from table \ref{tab:splitted_eval}, which is $85.9\,\%$. 
DA could reduce the average performance gap from $11.7\,\%$ to $7.4 \,\%$. 
Although this is still a rather large difference, DA could nevertheless compensate about one third of the original performance gap.


\begin{table}[!ht]
	\begin{center}
		\begin{tabular}{ c|c|c|c }
			Metric  & Before DA & After DA  & Improvement \\
			\hline
			\textit{OA}$[\%]$   &77.7& $81.7$ & $4.0$\\
			$\bar{F_1}$$[\%]$  &74.2& $78.5$ & $4.3$\\
			$\overline{IoU}$$[\%]$ &60.8& $65.9$ & $5.1$\\
			
		\end{tabular}
		\caption{\label{tab:adaptation-global} Average  global performance metrics before and after DA. }
	\end{center}
\end{table}

\begin{figure}[!ht]
	\begin{center}
		\begin{tabular}{m{0.3cm}c|m{1.6cm}m{1.6cm}m{1.6cm}m{1.6cm}m{1.6cm}}
			&& \multicolumn{1}{c}{$r^S$}&&&& \multicolumn{1}{c}{$r^T$} \\
			&$\mathcal{D}^S \rightarrow \mathcal{D}^T$ & \multicolumn{1}{c}{$y^S$}&\multicolumn{1}{c}{$x^S$}&\multicolumn{1}{c}{$x^{ST}$}&\multicolumn{1}{c}{$x^T$}&  \multicolumn{1}{c}{$y^T$}\\
			\hline
			1)&\textit{Hm}$\rightarrow$\textit{H}& 
			\vspace*{1mm} \includegraphics[width=1.6cm]{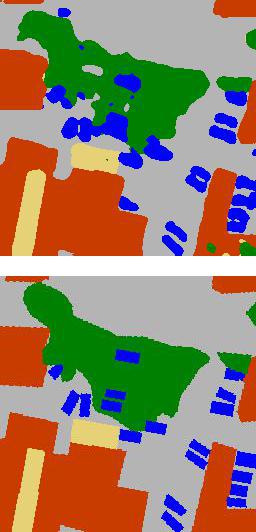}& 
			\vspace*{1mm} \includegraphics[width=1.6cm]{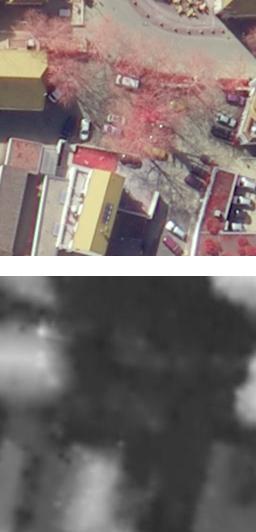}& 
			\vspace*{1mm} \includegraphics[width=1.6cm]{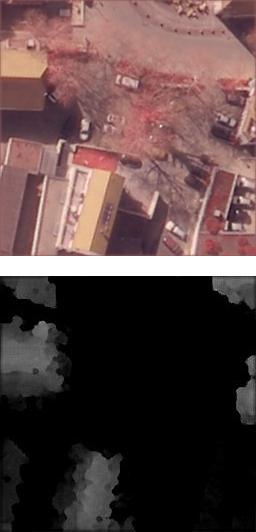}& 
			\vspace*{1mm} \includegraphics[width=1.6cm]{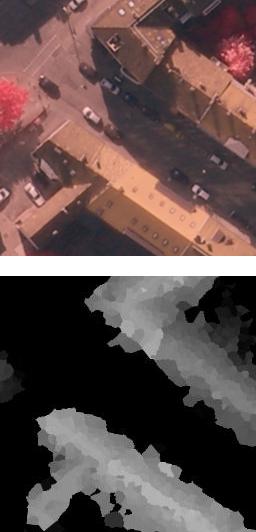}& 
			\vspace*{1mm} \includegraphics[width=1.6cm]{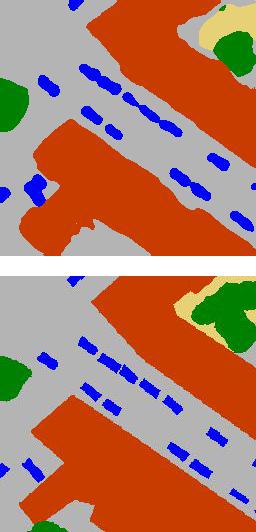} \\
			\cline{2-7}
			2)&\textit{H}$\rightarrow$\textit{Hm}& 
			\vspace*{1mm} \includegraphics[width=1.6cm]{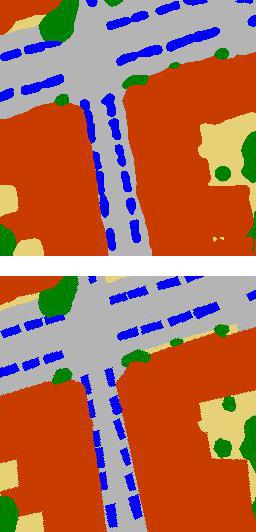}& 
			\vspace*{1mm} \includegraphics[width=1.6cm]{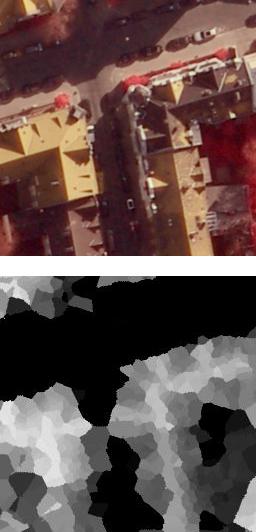}& 
			\vspace*{1mm} \includegraphics[width=1.6cm]{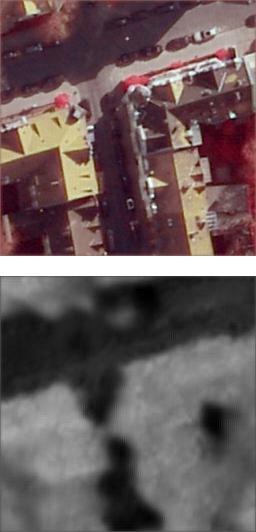}& 
			\vspace*{1mm} \includegraphics[width=1.6cm]{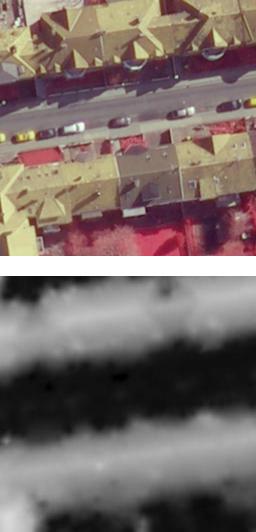}& 
			\vspace*{1mm} \includegraphics[width=1.6cm]{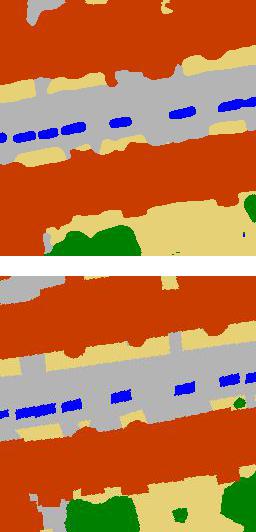} \\
			\cline{2-7}
			3)&$P'_{20}\rightarrow$\textit{N}& 
			\vspace*{1mm} \includegraphics[width=1.6cm]{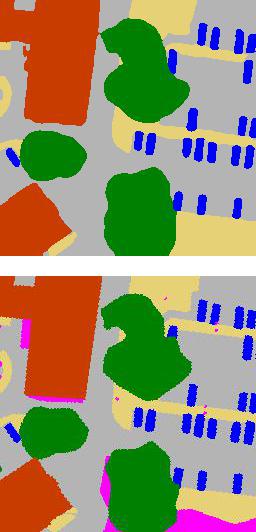}& 
			\vspace*{1mm} \includegraphics[width=1.6cm]{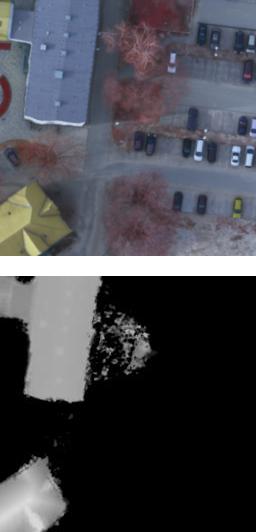}& 
			\vspace*{1mm} \includegraphics[width=1.6cm]{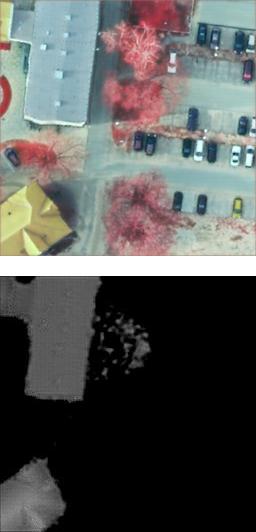}& 
			\vspace*{1mm} \includegraphics[width=1.6cm]{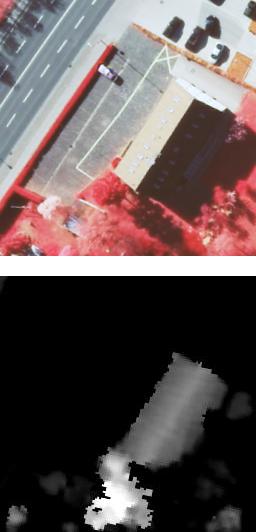}& 
			\vspace*{1mm} \includegraphics[width=1.6cm]{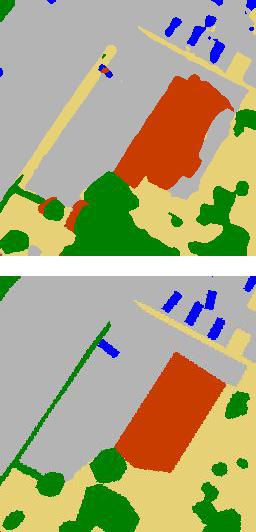} 	
		\end{tabular}
		\caption{\label{tab:examples} Qualitative examples of DA for three scenarios. Columns from left to right: Predicted (upper) and reference (lower) label map for $x^S$, image / nDSM $x^S$ from $\mathcal{D}^S$, transformed image / nDSM $x^{ST}$, example image / nDSM $x^{T}$, Predicted- (upper) and reference (lower) label map for $x^T$. Colour-codes: SG (grey), BU (red), HV (green), LV (yellow), VH (blue), CL (magenta). For the abbreviations of classes, cf.~table~\ref{tab:datasets}.}
	\end{center}
\end{figure}
\begin{figure}[ht]
	\begin{center}
		\begin{tabular}{m{0.3cm}c|m{1.6cm}m{1.6cm}m{1.6cm}m{1.6cm}m{1.6cm}}
			&& \multicolumn{1}{c}{$r^S$}&&&& \multicolumn{1}{c}{$r^T$} \\
			&$\mathcal{D}^S \rightarrow \mathcal{D}^T$ & \multicolumn{1}{c}{$y^S$}&\multicolumn{1}{c}{$x^S$}&\multicolumn{1}{c}{$x^{ST}$}&\multicolumn{1}{c}{$x^T$}&  \multicolumn{1}{c}{$y^T$}\\
			\hline
			4)&\textit{S}$\rightarrow$\textit{B}& 
			\vspace*{1mm} \includegraphics[width=1.6cm]{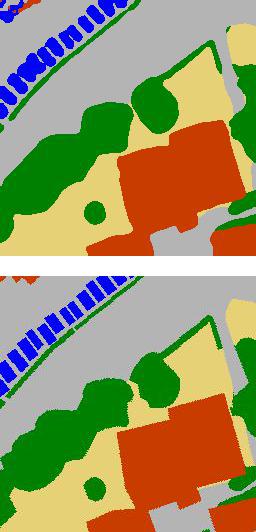}& 
			\vspace*{1mm} \includegraphics[width=1.6cm]{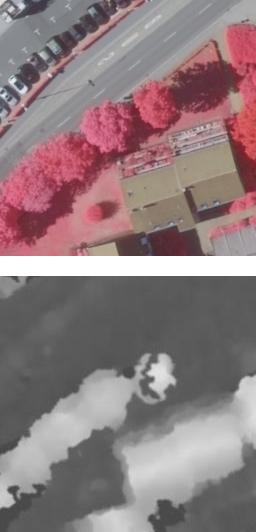}& 
			\vspace*{1mm} \includegraphics[width=1.6cm]{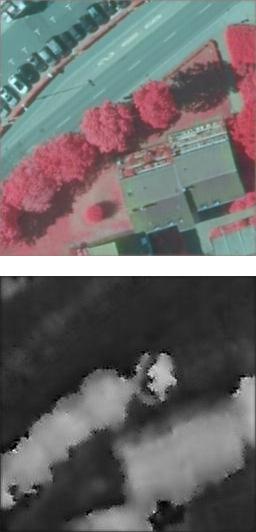}& 
			\vspace*{1mm} \includegraphics[width=1.6cm]{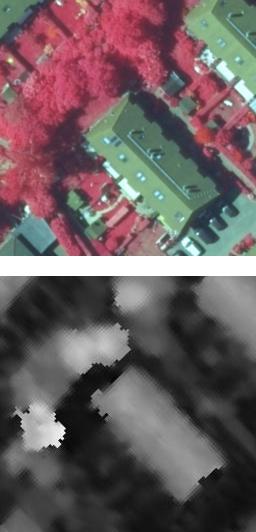}& 
			\vspace*{1mm} \includegraphics[width=1.6cm]{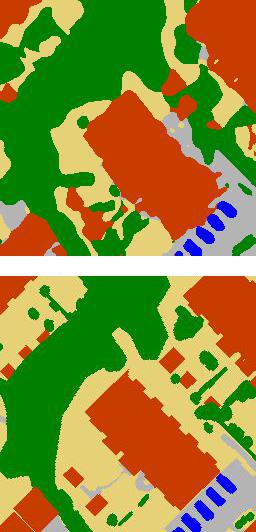} \\
			\cline{2-7}
			5)&\textit{B}$\rightarrow P'_{20}$& 
			\vspace*{1mm} \includegraphics[width=1.6cm]{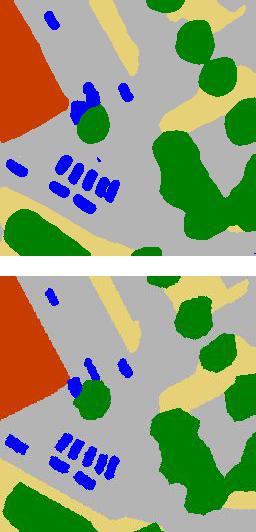}& 
			\vspace*{1mm} \includegraphics[width=1.6cm]{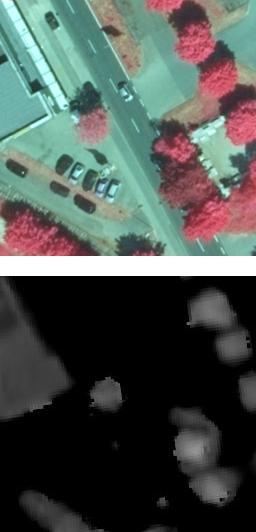}& 
			\vspace*{1mm} \includegraphics[width=1.6cm]{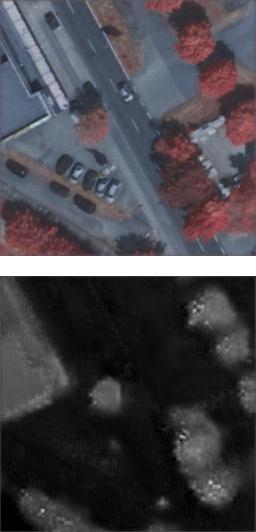}& 
			\vspace*{1mm} \includegraphics[width=1.6cm]{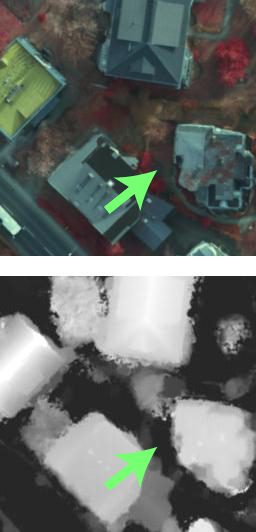}& 
			\vspace*{1mm} \includegraphics[width=1.6cm]{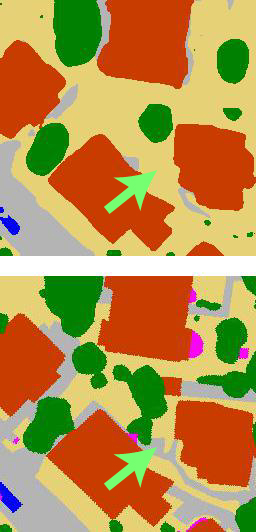} \\
			\cline{2-7}
			6)&\textit{N}$\rightarrow V'_{20}$& 
			\vspace*{1mm} \includegraphics[width=1.6cm]{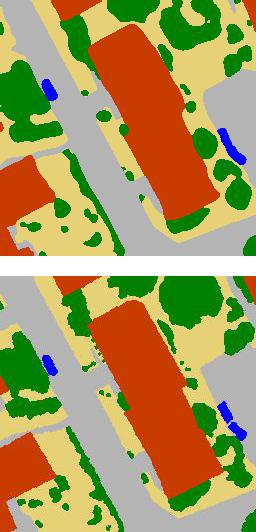}& 
			\vspace*{1mm} \includegraphics[width=1.6cm]{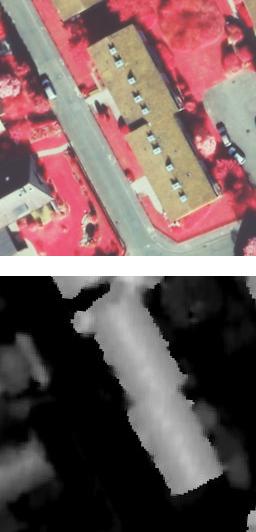}& 
			\vspace*{1mm} \includegraphics[width=1.6cm]{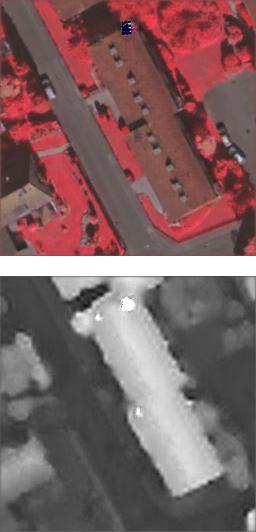}& 
			\vspace*{1mm} \includegraphics[width=1.6cm]{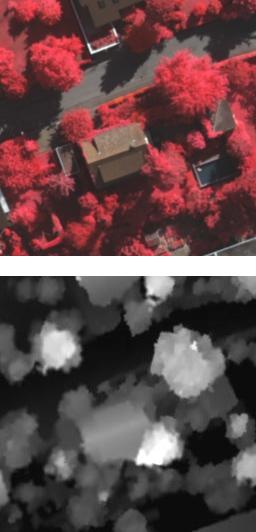}& 
			\vspace*{1mm} \includegraphics[width=1.6cm]{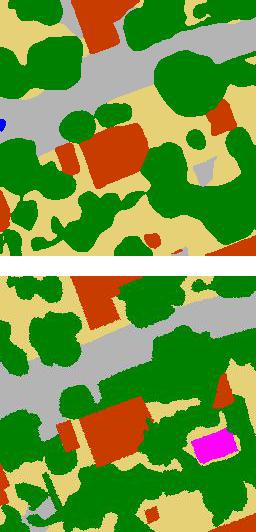} \\

		\end{tabular}
		\caption{\label{tab:examples2} More qualitative examples of the adaptation. Columns and rows as in figure \ref{tab:examples}. Green arrows in 5) indicate an example for a problematic region.}
	\end{center}
\end{figure}

Figures \ref{tab:examples} and \ref{tab:examples2} show some examples of appearance adaptation and the results of the classifier in $\mathcal{D}^T$.  
Note that the shown reference labels $y^T$ in $\mathcal{D}^T$ were  used neither for  training nor for DA. 
Having a look at the transformed images and nDSMs as well as exemplary samples from $\mathcal{D}^T$, it would seem that the style was adapted quite well. 
The examples 1) and 2) in figure~\ref{tab:examples}, involving the domain \textit{H}, should be highlighted. 
The nDSM is of relatively low quality in that domain, but nevertheless the figure indicates that in both cases, in which $H$ serves as target and source domain, respectively, style transfer works quite well for the height data.
A nice adaptation of the spectral channels can be seen in example 5) in figure \ref{tab:examples2}.
Whereas the image from  $\mathcal{D}^S$  is very bright and shows strong contrast, its transformed version is darker and has a lower contrast, corresponding to the appearance of the data in  $\mathcal{D}^T$. 
Some trees look rather greyish, which matches the appearance of the exemplary image from  $\mathcal{D}^T$. 


The predicted source domain labels shown in figures~\ref{tab:examples} and~\ref{tab:examples2}  are nearly error-free, which is to be expected because these images were used directly for training. 
There are more errors in the  predictions for the images from $\mathcal{D}^T$, where no reference labels were available for  training and DA. 
The majority of these errors is related to a confusion of high and low vegetation. 
These classes are actually hard to separate and the evaluation is highly influenced by the labelling policy, which is the reason why these problems of DA are not completely unexpected. 
In example 3) (figure~\ref{tab:examples}) the hedge was not detected as \textit{High Vegetation} in  $\mathcal{D}^T$, probably because no similar structure exists in  $\mathcal{D}^S$.
In examples of 4) and 5)  (figure~\ref{tab:examples2}), the large buildings were predicted rather well, but the small \textit{Building} instances are problematic. 
While in 4) their outlines are very imprecise, in 5) the small building in the centre of the patch was not predicted at all. 
In most examples the prediction of narrow paths belonging to \textit{Impervious Surface} is also problematic in $\mathcal{D}^T$.
Such a case is highlighted by the green arrows in example 5) (figure~\ref{tab:examples2}). 
However, such structures are often difficult to differentiate even for a human observer, particularly if the path is in a shaded area or if it is partially covered with dirt.


\subsection{Ablation studies}
\label{sec:ablation}

\subsubsection{Source training: adaptive loss}
\label{sec:sourcetraining}


In the first ablation study, we evaluate if the adaptive cross-entropy loss (ACE) proposed in section \ref{sec:ace} can help to improve the prediction quality of under-represented classes  based on the original Vaihingen dataset $V_9$ with the original class-structure. 
We compare the ACE loss to the regular cross-entropy loss (CE) and the multi-class focal loss (FCL). 
The models are trained according to the protocol described in section \ref{sec:setup} on the training area $\dot{V}_9$. 
Each experiment is repeated three times, each time starting from a different random initialization of the layers that are not pre-trained. 
Table \ref{tab:weighting} shows the means and standard-deviations of the achieved metrics using the provided reference label maps of $\tilde{V}_9$. 
For comparison to the leader-board of the ISPRS labelling benchmark, the version based on the ACE-loss is also evaluated on the eroded reference where the pixels close to the object boundaries are not considered (table \ref{tab:weighting}, last row).


%
%
%


\setlength{\tabcolsep}{0.25em}
\begin{table}[ht]
	\footnotesize
	\begin{center}
		\begin{tabular}{ c | c | c  c c c c c | c }
			\multicolumn{2}{c}{} & \multicolumn{6}{c}{class-wise F1-scores $[\%]$} & \\
			Loss & OA $[\%]$  & SG & BU & LV & HV & VH & CL & $\bar{F_1}$\\
			\hline
			\multicolumn{9}{c}{Full reference}\\
			\hline
			CE&$87.6\pm 0.0$&$89.5\pm 0.0$&$\textbf{94.0}\pm 0.2$&$80.0\pm 0.1$&$86.6\pm 0.2$&$80.3\pm 0.4$&$50.3\pm1.9$&\\ 
			FCL&$87.4\pm 0.1$&$89.1\pm 0.1$&$93.5\pm 0.1$&$79.8\pm 0.4$&$86.7\pm 0.1$&$76.7\pm 1.2$&$50.3\pm4.3$ &\\ 
			ACE&$\textbf{87.7}\pm 0.2$&$\textbf{89.6}\pm 0.1$&$93.7\pm 0.2$&$\textbf{80.7}\pm 0.3$&$\textbf{86.8}\pm 0.1$&$\textbf{81.7}\pm 0.5$&$\textbf{55.6}\pm 1.1$ &\\ 
			\hline
			\multicolumn{9}{c}{Eroded reference}\\
			\hline
       ACE&$90.6\pm 0.2$&$92.3\pm 0.1$&$95.5\pm 0.2$&$84.1\pm 0.4$&$89.8\pm 0.1$&$88.2\pm 0.4$& $58.3\pm 0.8$ & \\ 
			\hline
		\end{tabular}
		\caption{\label{tab:weighting} Results on the Vaihingen dataset from the ISPRS semantic labelling benchmark. Scores show mean and standard-deviation over three runs. Best results when comparing to the full reference label maps are printed in bold font. Last row:  metrics achieved when using the eroded reference for evaluation. }
	\end{center}
\end{table}


The table indicates that the best performance is shown by the model based on the ACE loss. 
It does not only achieve the best OA and $\bar{F_1}$, but also the highest class-wise F1-scores for five out of six classes; for the sixth class the difference to the best model is below 0.3\%. 
In general, the variation in OA is very low: the difference between the best model and one achieving the worst OA, the model trained from scratch using the focal loss, is smaller than 0.5\%. 
The largest impact of the loss can be observed in the F1-score of the under-represented classes \textit{Vehicle} and \textit{Clutter}. 
The proposed ACE-loss achieves the highest metrics for these classes; in case of \textit{Clutter}, the improvement is about 5\%. 
Compared to the regular CE loss, the focal loss resulted in a reduced performance for \textit{Vehicle} and had no impact on the F1-score of \textit{Clutter}. 
The  metrics achieved on the eroded reference are comparable to the best-performing results in the leader-board of the benchmark. 
We want to point out that according to the leader-board\footnote{www2.isprs.org/commissions/comm2/wg4/results/} there is only one listed approach that achieves higher values for the F1-score of the class \textit{Vehicle}, unfortunately without a publication.
The variant using  the proposed ACE loss leads to a $\bar{F_1}$ value of $80.9\,\%$. 
This is comparable to the best result in \citep{yang2020} on this benchmark ($80.8\,\%$), where cosine similarity and the focal loss were used to address class imbalance.

\begin{figure}[!ht]
	\centering
	\includegraphics[width=.5\textwidth]{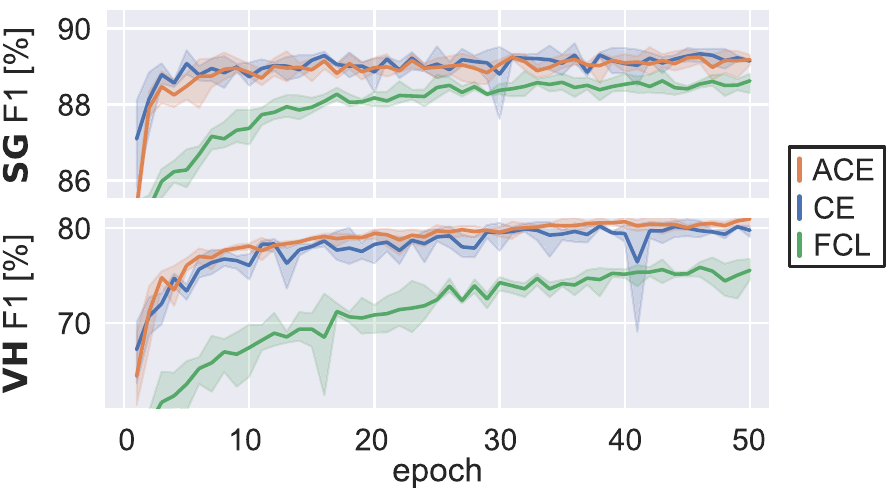}
	\caption{Mean of class-wise F1-scores for the frequent class \textit{Sealed Ground} (SG) and the under-represented class \textit{Vehicle} (VH), averaged over three runs. The shaded areas correspond to the 95\% confidence intervals.}\label{fig:supervised_V}
\end{figure}

In order to analyse the performance during training and the convergence behaviour, the class-wise F1-scores of two classes on the test set were tracked during training (cf. figure \ref{fig:supervised_V}). 
The scores are based on the non-eroded reference, and no horizontal and vertical flipping was used during inference, which is the reason why the results from the last epoch are slightly worse than those reported in table~\ref{tab:weighting}. 
As far as the under-represented class \textit{Vehicle} is concerned, figure \ref{fig:supervised_V} shows that the ACE loss outperforms the other losses and the focal loss performs worst by a relatively large margin.
For the frequent class \textit{Sealed Ground} there is no clear difference between using CE and ACE, but both also outperform the focal loss.
To summarize, the proposed ACE loss mitigates the problems of imbalanced class distributions to some degree compared to the other loss functions. 


\subsubsection{Influence of the regularization term for the discriminator}
\label{sec:tuning}


In this section we analyse the influence of the hyper-parameter $\rho$ that controls the influence of the regularization loss $\mathcal{L}_{reg}$ of the discriminator (cf. equation~\ref{eq:Ldiss}) on the performance of DA. 
To that end, the adaptation from \textit{S} to \textit{Hm}, which was the scenario used to tune the hyper-parameters, is performed several times using different values for $\rho$ in the range between $\rho = 0$, which  implies the regularization loss is not considered, to $\rho = 8$. 
The experiment is repeated three times for each value. 
In table \ref{tab:tuning} the means and the standard deviations for several quality metrics after adaptation are given for each tested value of $\rho$.


\begin{table}[ht]
	\begin{center}
		\begin{tabular}{ c | c c c |c |c|}
			$\rho$ & $OA$ & $\bar{F_1}$ & $\overline{IoU}$ & $PTR(\bar{F_1})$ & $PTR(OA)$\\
			\hline
			0 & $81.1\pm 1.2$ & $77.6\pm 1.1$ & $64.0\pm 1.4$ & $1/3$ & $1/3$ \\ 
			\hline
			1 & $81.6\pm 1.2$ & $78.7\pm 0.9$ & $65.5\pm 1.2$ & $2/3$ & $2/3$ \\ 
			\hline
			2 & $81.7\pm 0.3$ & $79.0\pm 0.2$ & $65.9\pm 0.3$ & $3/3$ & $3/3$\\ 
			\hline
			4 & $\textbf{82.3}\pm 0.3$ & $\textbf{79.3}\pm 0.1$ & $\textbf{66.2}\pm 0.2$ & $3/3$ & $3/3$\\ 
			\hline
			6 & $80.4\pm 0.5$ & $77.4\pm 0.3$ & $63.7\pm 0.4$ & $0/3$ & $0/3$\\ 
			\hline
			8 & $78.2\pm 1.6$ & $75.3\pm 1.7$ & $61.0\pm 2.1$ & $0/3$ & $0/3$\\ 
			\hline
			ST & $81.8$ & $78.5$ & $65.2$ & \multicolumn{2}{c}{} 
		\end{tabular}
		\caption{\label{tab:tuning} Influence of the hyper-parameter $\rho$ on the adaptation scenario $\textit{S}\rightarrow \textit{Hm}$. Best results are printed in bold font. $\rho = 0$ corresponds to training without the regularization loss. ST: source training without DA. $PTR(\bar{F_1})$,  $PTR(OA)$:  number of positive transfers with respect to $\bar{F_1}$ and OA, respectively. }
	\end{center}
\end{table}


Among all tested values for $\rho$, DA performs best for $\rho=4$, where a positive transfer with respect to $\bar{F}_1$ and $OA$ is observed in all three test runs. 
The latter statement also applies for $\rho=2$, but in this case, the resulting scores are worse. 
Increasing $\rho$ beyond 4 as well as decreasing it results in a worse DA performance. 
Disabling the auxiliary loss by setting $\rho=0$ resulted  in a positive transfer only once with respect to all of the listed global metrics. 
Whereas the improvement in OA and  $\bar{F}_1$  due to $\mathcal{L}_{reg}$ is very small (1.2\% and 1.7\%, respectively), we conclude that this regularization loss has a stabilizing influence on DA and helps to avoid negative transfer if the hyper-parameter  $\rho$ is tuned properly.

\begin{figure}[!ht]
	\centering
	\includegraphics[width=0.9\textwidth]{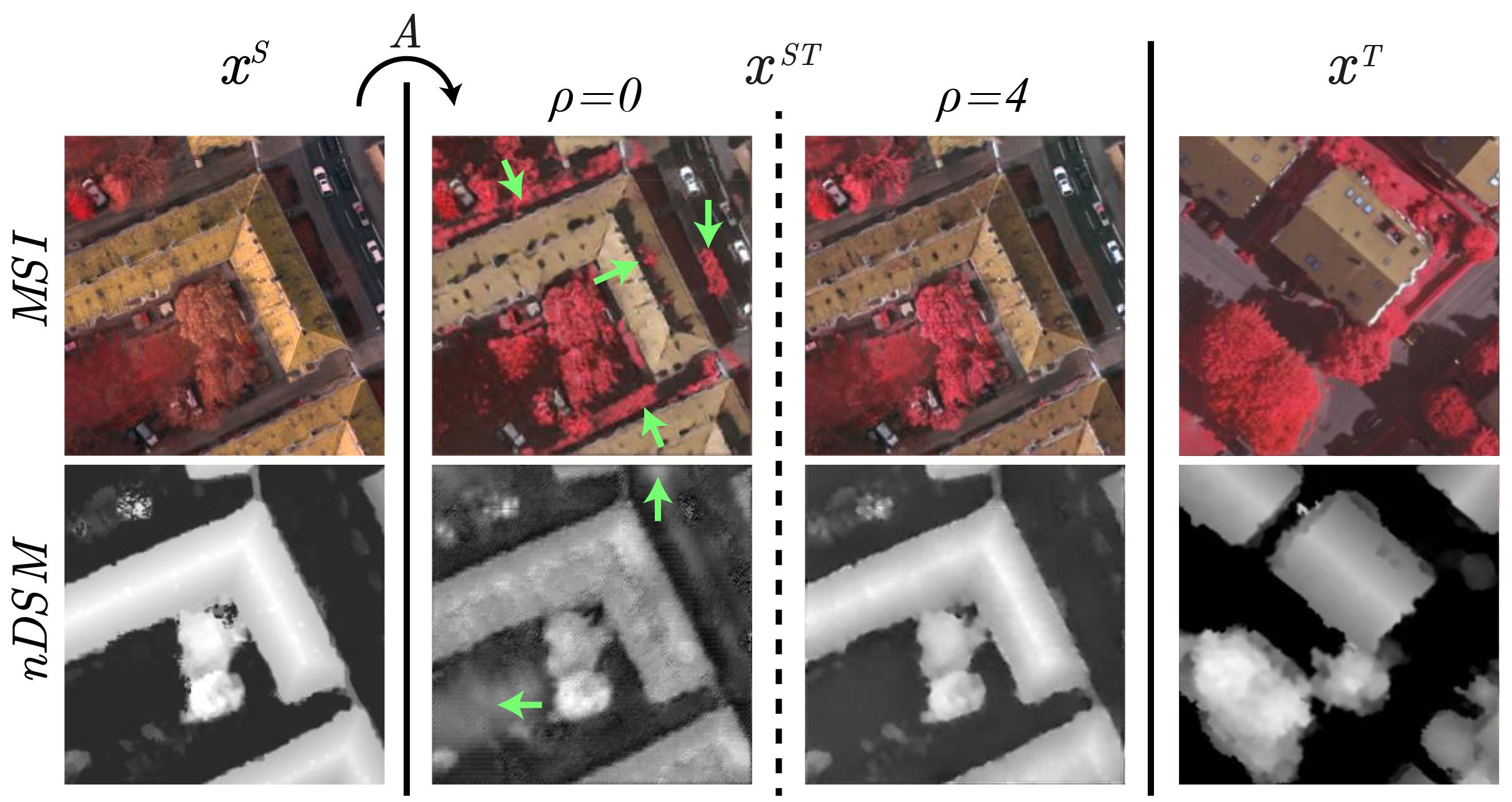}
	\caption{Visual comparison for the appearance adaptation from $P_{20}$ to $V_{20}$ when using $\rho=0$ (centre left) and $\rho=4$ (centre right). Green arrows: hallucinated structures in the transformed data. Right column: example from $\mathcal{D}^T$.}\label{fig:rho_example}
\end{figure}

Figure~\ref{fig:rho_example} shows a visual example of the appearance adaptation from $P_{20}$ to $V_{20}$ with and without the regularization loss.  
This is a difficult scenario because trees occur more frequently in $V$ and they also have a different appearance due to seasonal effects.
The model which was trained using $\rho = 4$ delivers reasonable appearance adaptation results, whereas the model which was trained without the proposed regularization leads to artifacts in the transformed image and nDSM; some of them are highlighted by green arrows in figure~\ref{fig:rho_example}.
Without regularization the model hallucinates structures with a large reflectance in the infrared band, most likely due to the higher occurrence of such regions in the target domain.


\subsubsection{Evaluation of the parameter selection criterion}
\label{sec:discussion_stopping}

\begin{figure}[!ht]
	\centering
	\includegraphics[width=0.9\textwidth]{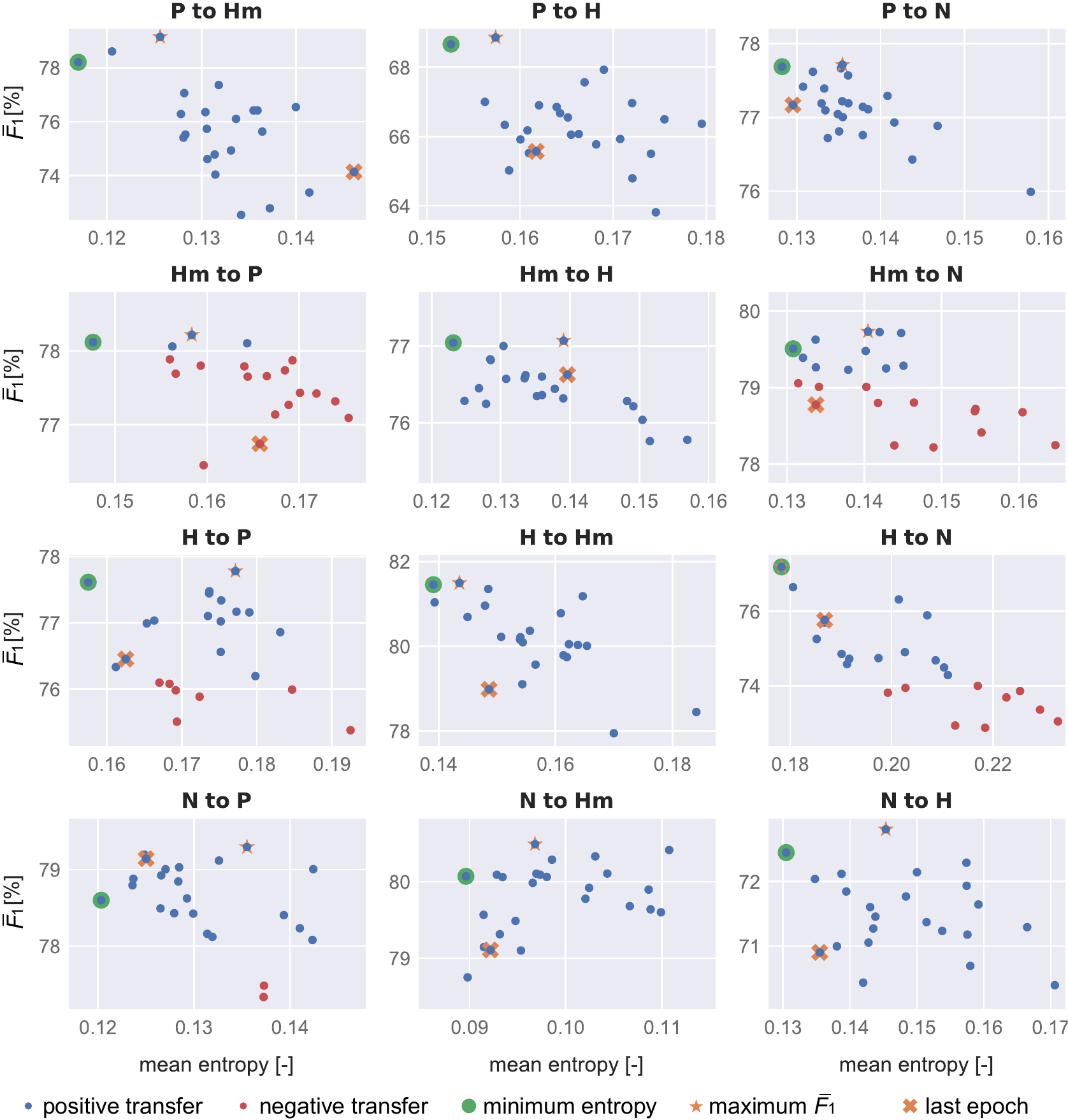}
	\caption{Visualisation of the mean F1 scores achieved in $\mathcal{D}^T$ as a function of the mean entropy in the DA epochs 26-50 for several DA scenarios. Each point corresponds to one DA epoch. The heading of each graph encodes the scenario ($\mathcal{D}^S$ to $\mathcal{D}^T$). Other DA scenarios resulted in  similar graphs, but are omitted to save space. }\label{fig:entro_influence}
\end{figure}

The entropy-based criterion for parameter selection proposed in section~\ref{sec:eps}  
is based on the assumption that parameter sets which achieve a lower mean entropy in $\mathcal{D}^T$ also have a better performance in that domain. 
To validate this assumption, the $\bar{F}_1$ values were tracked for epochs 26-50 in the DA experiments described in section~\ref{sec:eval_of_da}, so that $\bar{F}_1$  can be analysed  as a function of the mean entropy in $\mathcal{D}^T$. 
The results are shown in figure \ref{fig:entro_influence} for DA scenarios involving four domains (two autumn domains - $P$, $Hm$ - and two summer domains - $H$, $N$). 
Every point corresponds to one epoch in DA training and, thus, to a set of model parameters values; 
parameter sets which resulted in a positive transfer are shown in blue, those that yielded a negative transfer in red. 
Furthermore, three parameter sets are highlighted: the set after the last epoch (red cross), the one resulting in the highest mean F1-score (star), and the set achieving the minimum entropy, i.e. the set selected by our method (large green dot). 


Figure \ref{fig:entro_influence} shows that in many scenarios the performance is higher for parameter sets that also achieved a low entropy in $\mathcal{D}^T$. 
Whereas the parameters chosen according to our selection criterion, i.e. those with the lowest entropy (green dots), correspond to a positive transfer in all 42 DA scenarios, the parameters after the last epoch lead to a negative transfer in three cases. 
In 26 out of the 42 scenarios the parameter sets with minimum entropy resulted in a better performance than those obtained after the last training epoch, in 11 the results were slightly worse and in 5 cases they were almost identical. 
On average, using the model from the last training epoch results in a $\bar{F}_1$  value of $76.6\,\%$, while using the proposed criterion results in $\bar{F}_1 = 78.0\,\%$. 
This is rather close to the average $\bar{F}_1$  value of the best models ($78.4\,\%$).
Particularly in difficult scenarios such as $Hm \rightarrow P$ or $Hm \rightarrow N$, where many parameter sets resulted in a negative transfer, the set selected according to our entropy-based criterion achieved a positive transfer. 
Furthermore, the improvement was quite large in a few cases. 
For example, in $P'_{20} \rightarrow Hm$ and $V'_{20} \rightarrow H$ the improvement is around $4\,\%$ w.r.t $\bar{F}_1$.
The largest difference can be observed in the scenario $S \rightarrow H$ where the last model ($42.4\,\%$) has a much lower mean F1-score than the model which is selected according to the proposed criterion ($65.4\,\%$).
In the worst-case scenario $B \rightarrow P'_{20}$ the last model is only $0.6\,\%$ better than the selected one.
We take this as an indication that our selection criterion is better than picking the last model. 
It must be noted that the overall maximum number of 50 epochs was chosen empirically, and there might be another fixed number of epochs that would achieve better results, but this would require a rather difficult tuning procedure based on multiple scenarios. 
We thus conclude that, whereas our selection criterion might not result in the best choice of parameters in all scenarios, it achieves a consistently good choice. 
Although on average, the improvement in $\bar{F}_1$ is relatively small ($1.4\,\%$), the selection criterion definitively helps to avoid negative transfer, achieving a positive transfer in all investigated scenarios, and in some cases can lead to a significant improvement.


\subsection{Comparison to the state of the art}
\label{sec:comparative}


\subsubsection{Comparison to instance transfer}\label{sec:comp-entmin}


In this section we compare the proposed method to the variant of instance transfer described in \citep{wittich20entmin}. 
For this comparison we use the same inference protocol and the same datasets $O\in\{P'_{20},V'_{20},S,Hm,B\}$ with the same split into training and test sets that were also used in that publication. 
In the initial cross-domain evaluation before DA, models trained using the training data from $\mathcal{D}^S$ are evaluated on the test sets of all other domains. 
All models are then adapted to the other domains, using the labelled training samples from $\mathcal{D}^S$ and all unlabelled samples from $\mathcal{D}^T$, and the results are evaluated using the test samples in $\mathcal{D}^T$. 
The results of the classifier trained on the training sets $\dot{O}$ and applied to the test sets of the same domains $\tilde{O}$, indicating optimal case  of classification, are those already presented in section~\ref{sec:adaptation-1-same}. 
The results achieved for instance transfer are those reported in \citep{wittich20entmin} for the same test setup. 
For the comparison, we focus on $\bar{F}_1$, averaged over all 20 possible combinations of source and target domains. 
The obtained average metrics using both methods are given in table~\ref{tab:entmin}.

\begin{table}[H]
	\begin{center}
		\begin{tabular}{r | c | c| c | c }
			& \multicolumn{2}{c|}{\citep{wittich20entmin}} & \multicolumn{2}{c}{this work}\\
			& $\bar{F_1}\,[\%]$ & $OA\,[\%]$ & $\bar{F_1}\,[\%]$ & $OA\,[\%]$\\
			\hline
			$\mathcal{D}^S\rightarrow\mathcal{D}^S$& $83.6$ &$87.9$& $85.9$&$88.4$\\
			$\mathcal{D}^S\rightarrow\mathcal{D}^T$ (no DA) & $72.1$ &$77.5$& $75.4$&$78.9$\\
			$\mathcal{D}^S\rightarrow\mathcal{D}^T$ (after DA) & $74.7$ &$79.3$& $78.1$&$81.5$\\
			Improvement due to DA & $2.6$ &$1.8$& $2.7$&$2.5$\\
			\hline
			Remaining gap & $8.9$ &$8.6$& $7.8$&$6.9$\\
			
		\end{tabular}
		\caption{\label{tab:entmin} Comparison to \citep{wittich20entmin}. The table shows $OA$ and $\bar{F_1}$ averaged over all combinations of source and target domains. The improvement due to DA is the difference between $OA$ and $\bar{F_1}$ scores in $\mathcal{D}^T$ after and before DA, respectively. The remaining gap is the difference between the optimal $OA$ and $\bar{F_1}$ scores ($\mathcal{D}^S\rightarrow\mathcal{D}^S$) and those achieved after DA. }
	\end{center}
\end{table}


The approach proposed in this paper outperforms the method from \citep{wittich20entmin} in all DA scenarios. 
Although the absolute improvement due to DA is only slightly increased ($2.7\,\%$ vs. $2.6\,\%$), the results of the proposed method before DA are already at the level achieved by \citep{wittich20entmin} after DA. 
The proposed source training strategy obviously leads to an improved generalization of the models in both, intra- and cross-domain scenarios, which may be due to the usage of the pre-trained backbone that was not used in \citep{wittich20entmin}. 
The value of $\bar{F_1}$  after DA is better by 4.3\% than the one achieved in \citep{wittich20entmin}. 
Having a look at the remaining gap between the performance achieved in the intra-domain settings and the  performance after DA, it can be seen that it was further reduced by about 1\% in $\bar{F_1}$  compared to our previous work.
However, there is still a gap of almost $8\,\%$ left.


\subsubsection{Comparison to other DA methods} \label{sec:comp-multi}

In this section, the proposed method is compared to three DA approaches 
that were evaluated on the basis of the data of the ISPRS Semantic Labelling challenge, i.e., $P$ and $V$, using one of them as $\mathcal{D}^S$  and the other as $\mathcal{D}^T$  \citep{liu19curve,benjdira2019unsupervised,ji2020fullspace}. 
The methods were already briefly described in section~\ref{sec:relwork-DA}.  
As the  benchmark was not designed for comparing DA approaches, the authors of these  papers used different protocols for evaluating their methods. 
We adapt our test setup to the ones used by the authors of the methods we compare to, which leads to three different evaluation scenarios. 
In each case, we conduct one experiment in which we use the same input channels as the approaches we compare to. 
As the authors of the original papers did not use height data, this means that  we do not use the nDSM in these experiments, either. 
In a second set of experiments, we include the height information to evaluate its impact on DA. 
A major difference of our approach to those we compare to is that we compensate for differences in the resolutions (cf.~section \ref{sec:resolutiontransfer}). 
In any case, the actual evaluation is always based on the same dataset (and at the same resolution) as in the original papers. 
In the following, we briefly describe the three evaluation protocols before discussing their results.


\paragraph{\bf I) \citep{liu19curve}} This method was evaluated by adapting from the subset $\bar{V}'_9$  to $P'_5$. 
\citet{liu19curve} only use the \texttt{IRG} channels at the original GSDs for both domains, and they ignore the class \textit{Clutter} both in training and evaluation. 
In our experiments we first train a model on $\bar{V}'_9$ and then adapt it to $P'_9$ before upsampling the predictions to the GSD of $5\,cm$ of $\mathcal{D}^T$. 
In the first variant (I$_a$) we only use the \texttt{IRG} data, in the second variant (I$_b$) the nDSM is also used. 


\paragraph{\bf II) \citep{benjdira2019unsupervised}} 
This method was evaluated by transferring from $P_5$ to $V_9$. 
\citet{benjdira2019unsupervised} use the $\mathtt{RGB}$ data in $P_5$ and $\mathtt{IRG}$ in $V_9$, which violates the definition of DA according to \citep{tuia2016} because $\mathcal{X}^S \neq \mathcal{X}^T$, making this a very difficult task. 
We follow our training protocol, downsampling the Potsdam data to $P_{9}$ to be able to use it as $\mathcal{D}^S$, and using $V_{9}$ as $\mathcal{D}^T$. 
Like \citet{benjdira2019unsupervised}, we also evaluate a variant of our method where we use the $\mathtt{RGB}$ data is input space of $\mathcal{D}^S$ and transfer to $V$, where we have  $\mathtt{IRG}$  data (variant II$_a$). 
The second variant (II$_b$) uses $\mathtt{RGBH}$ as feature space for $\mathcal{D}^S$ and $\mathtt{IRGH}$ in $\mathcal{D}^T$.
In both abbreviations, $\mathtt{H}$ denotes the nDSM.


\paragraph{\bf III) \citep{ji2020fullspace}} 
This method was evaluated by transferring from $P_5$ to $V_9$, downsampling the Potsdam data to a GSD of $10\,cm$ before adapting to $V_9$. 
This is very similar to our strategy for dealing with different resolutions, but we downsample directly to the target GSD of $9\,cm$. 
The first variant of our experiment (III$_a$) uses the  \texttt{IRG} channels for DA, which is the setting of \citep{ji2020fullspace}. 
The second variant (III$_b$) includes the nDSM as an additional channel. 


\paragraph{\bf Results} The results of the experiments for comparing our method to the three methods described above are presented in table \ref{tab:comparison}. 
For a first assessment of the results we compare the results of our method when using the same input features as the baseline methods (the variants with index $a$). 
In all cases we outperformed the methods we compare to by quite a  large margin with respect to the global performance metrics, both after source training and after DA.
Interestingly, we achieve better results without DA than the baseline methods do with DA. 
In I) and II) this is not unexpected because unlike the corresponding methods we compare to we adapt the GSDs as described in section~\ref{sec:resolutiontransfer}.  
We take this as an indication that the adaptation of the resolution by resampling should be preferred over DA without such a pre-adaptation.
In III), however, the authors use the resampled version $P_{10}$ as source domain, so that the remaining difference in the GSD should not have a large impact. 
Nevertheless, our source-trained model still performs better than method III does after DA, which we believe to be due to the use of strong data augmentation, which can improve the cross-domain performance of the models before adaptation considerably \citep{wittich20entmin}. 
In case of III), we can also compare the class-wise performance metrics. 
Their behaviour is rather similar except for the class LV, for which our DA method resulted in a negative transfer and \citep{ji2020fullspace} performed better. 


\setlength{\tabcolsep}{0.3em}
\begin{table}[ht]
	\footnotesize
	\begin{center}
		\begin{tabular}{ c c |  c | c  c  c | c c c c c c | c }
			
			\multicolumn{3}{c}{}&\multicolumn{3}{c|}{$[\%]$}  & \multicolumn{6}{c}{class-wise $IoU\,[\%]$} &\\
			& $\mathcal{X}^S  , \mathcal{X}^T $& DA & $OA$ & $\bar{F}_1$ & $\overline{IoU}$ &SG & BU & LV & HV & VH & CL&Var. \\
			\hline
			\hline
			\multicolumn{12}{c|}{\multirow{2}{*}{$\bar{V}'_9 \rightarrow P'_5$}}&\\
			\multicolumn{12}{c|}{ }&\\
			\hline
			\multirow{2}{*}{\textbf{I)}}&\multirow{2}{*}{\rotatebox[origin=c]{90}{\parbox[c]{0.7cm}{\centering $\mathtt{IRG}$}}}
			& N                   & $50.6$ & \multirow{2}{*}{n.r.} & $32.3$ &\multicolumn{5}{c}{\multirow{2}{*}{n.r.}}& -& \multirow{2}{*}{}\\ 
			&& Y           & $62.1$ &  & $38.7$  &&&&&& - &\\ 
			\hline
			\multirow{4}{*}{\rotatebox[origin=c]{90}{\parbox[c]{1.5cm}{\centering \textbf{ours}}}} 
			&\multirow{2}{*}{\rotatebox[origin=c]{90}{\parbox[c]{0.7cm}{\centering $\mathtt{IRG}$}}}
			&   N    & $68.5$ & $67.4$ &$53.2$& $53.0$ & $68.0$ & $49.9$ & $21.1$ & $74.2$ & - & \multirow{2}{*}{I$_a$}\\ 
			&& Y & $72.5$ & $71.2$ &$57.5$&$60.6$ & $73.7$ & $52.9$ & $26.0$ & $74.5$ & - &\\
			\cline{2-13}
			& \multirow{2}{*}{\rotatebox[origin=c]{90}{\parbox[c]{0.7cm}{\centering $\mathtt{IRGH}$}}}
			&   N    & $74.2$ & $72.6$ &$58.8$& $61.1$ & $77.9$ & $54.7$ & $30.5$ & $69.7$ & - & \multirow{2}{*}{I$_b$}\\ 
			&& Y & $76.0$ & $74.3$ &$61.1$& $65.3$ & $79.5$ & $56.1$ & $31.1$ & $73.6$ & - &\\
			\hline
			\hline
			\multicolumn{12}{c|}{\multirow{2}{*}{$P_5 \rightarrow V_9$}}\\
			\multicolumn{12}{c|}{ }&\\
			\hline
			\multirow{2}{*}{\textbf{II)}}  &\multirow{2}{*}{\rotatebox[origin=c]{90}{\parbox[c]{0.7cm}{\centering $\mathtt{RGB}$\\$\mathtt{IRG}$}}}
			& N                   & \multirow{2}{*}{n.r.} & $32$ & $17$ &\multicolumn{6}{c|}{\multirow{2}{*}{n.r.}}& \multirow{2}{*}{}\\ 
			&& Y           &  & $49$ & $30$  &&&&&& &\\ 
			\hline
			\multirow{4}{*}{\rotatebox[origin=c]{90}{\parbox[c]{1.5cm}{\centering \textbf{ours}}}} 
			& \multirow{2}{*}{\rotatebox[origin=c]{90}{\parbox[c]{0.7cm}{\centering $\mathtt{RGB}$\\$\mathtt{IRG}$}}}
			&   N       & $56.2$ & $50.2$ &$36.7$&$35.1$&$56.7$&$18.6$&$55.2$&$51.2$&$3.8$ &\multirow{2}{*}{II$_a$}\\ 
			&&  Y       & $75.6$ &$66.9$ &$53.5$&$63.9$&$77.5$&$43.8$&$64.1$&$59.2$&$12.7$\\
			\cline{2-13}	
			& \multirow{2}{*}{\rotatebox[origin=c]{90}{\parbox[c]{0.7cm}{\centering $\mathtt{RGBH}$\\$\mathtt{IRGH}$}}}
			&   N       &$75.3$&$64.9$&$52.8$&$69.7$&$82.8$&$34.0$&$65.6$&$56.9$&$7.8$&\multirow{2}{*}{II$_b$}\\ 
			&& Y        &$78.8$&$68.1$&$56.2$&$73.5$&$85.7$&$40.8$&$68.0$&$59.3$&$10.1$&\\
			\hline
			\hline
			\multirow{2}{*}{\textbf{III)}}&\multirow{2}{*}{\rotatebox[origin=c]{90}{\parbox[c]{0.7cm}{\centering $\mathtt{IRG}$}}}		
			  & N & $52.6$ &\multirow{2}{*}{n.r.}& $26.7$ &$34.2$ &$44.1$ &$16.7$ &$53.3$ &$7.5$ & $4.3$&\multirow{2}{*}{} \\ 
			& & Y                   & $68.2$ && $43.7$ &$57.4$ &$57.8$ &$41.7$ &$58.4$ &$37.0$ & $10.0$ \\ 
			\hline
			\multirow{4}{*}{\rotatebox[origin=c]{90}{\parbox[c]{1.5cm}{\centering \textbf{ours}}}} 
			& \multirow{2}{*}{\rotatebox[origin=c]{90}{\parbox[c]{0.7cm}{\centering $\mathtt{IRG}$}}}
			&  N       & $73.7$ & $64.1$&$50.1$&$57.4$&$71.8$&$43.3$&$66.1$&$49.7$ &$12.2$&\multirow{2}{*}{III$_a$}\\ 
			&& Y        & $76.5$ & $67.2$ &$53.9$& $67.0$ & $77.0$ & $40.5$ & $65.5$ & $58.9$ & $14.3$ &\\
			\cline{2-13}
			
			& \multirow{2}{*}{\rotatebox[origin=c]{90}{\parbox[c]{0.7cm}{\centering $\mathtt{IRGH}$}}}
			
			&   N  & $74.5$ & $65.2$ &$53.4$& $70.9$ & $84.4$ & $38.2$ & $65.8$ & $55.5$ & $5.7$ &\multirow{2}{*}{III$_b$}\\ 
			&&  Y  & $79.4$ & $69.8$ &$57.3$& $70.3$ & $82.8$ & $46.7$ & $68.5$ & $61.2$ & $14.0$ &\\
			
		\end{tabular}
		\caption{Comparison to I) \citep{liu19curve}, II) \citep{benjdira2019unsupervised} and III) \citep{ji2020fullspace} based on DA between Potsdam and Vaihingen. Column DA indicates whether the results were achieved with DA (Y) or without DA (N), the latter case indicating that the source classifier was applied to  $\mathcal{D}^T$ without any adaptation. 
			Var.: variant of our method. 
			n.r.: not reported in the original publication. 
			\label{tab:comparison}}
	\end{center}
\end{table}


\begin{figure}[!b]
	\centering
	\includegraphics[width=0.6\textwidth]{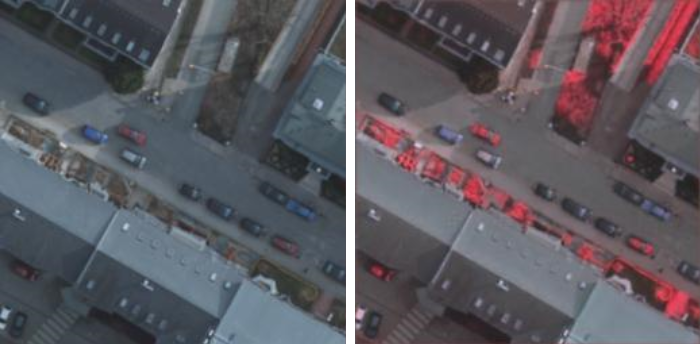}
	\caption{Example for the appearance adaptation from \texttt{RGB} (left) to \texttt{IRG} (right) ($P$).}\label{fig:rgb-irg-example}
\end{figure}

Looking at the results of experiment II$_a$, which involves an adaptation from  \texttt{RGB} in $\mathcal{D}^S$ to \texttt{IRG} in $\mathcal{D}^T$, we note a surprisingly good performance of our DA approach. 
The poor performance of the classifier trained in $\mathcal{D}^S$ when being applied to data from $\mathcal{D}^T$ without DA is not surprising. 
Our DA approach improves OA by almost 20\% and $\bar{F}_1$ by about 16\%, and the metrics achieved are in the same range of values as those in the other scenarios (where we also use \texttt{IRG} in $\mathcal{D}^S$). 
Although the appearance adaptation network was not specifically designed to transform between different input spaces, it would seem that it is quite successful under these circumstances nevertheless. 
A qualitative example for such a transfer is shown in figure \ref{fig:rgb-irg-example}.


The results for variant II$_b$ show that including height information does increase the performance in $\mathcal{D}^T$ before and after adaptation even when the input spaces for $\mathcal{D}^S$ and $\mathcal{D}^T$ are different. 
However, using \texttt{IRG} as feature space in both domains (III$_a$) gives better metrics, which was to be expected because the domain difference is much smaller. 
The best results are obtained when further including height information (III$_b$). 
In all cases our proposed DA strategy improved the global quality metrics and in most cases also the class-wise metrics.
We note that the performance for the class \textit{Clutter} of the adapted models is rather poor in all cases, which we think is because the semantic meaning of this class is different in $V$ and $P$. 



\section{Summary and Outlook}
\label{sec:summary}


In this work, we have presented a method for semi-supervised DA for  the pixel-wise classification of images and, potentially, height data that is based on adversarial appearance adaptation.  
Unlike existing work, it does not rely on cycle-consistency or cross-cycle-consistency to achieve semantic consistency in the appearance adaptation, but it does so by joint training of the appearance adaptation and the classification networks. 
A new regularization term for the domain discriminator was designed to mitigate problems due to different label distributions in the source and target domains.
We also proposed a new parameter selection criterion, which at least partially solves the problem of missing validation data in the target domain. 
Finally, we developed a new variant of a weighted loss for supervised training with imbalanced training datasets. 


Our experiments indicate that the new weighted loss can indeed improve the performance of the resultant classifier for under-represented classes, though there still remains a performance gap compared to the more frequent classes. 
The proposed regularization term of the discriminator was found to be beneficial for the success of DA, stabilizing it so that a positive transfer could be achieved in all compared scenarios if the corresponding hyper-parameter was tuned properly. 
In our experiments for the evaluation of our DA method we could show that our method leads to a positive transfer in all tested scenarios. 
The performance gain due to DA compared to applying the classifier trained on source domain data to the target domain without adaptation are quite satisfactory with an average improvement of $4.0\,\%$ in  OA and $4.3\,\%$ in the mean F1-score, averaged over 41 DA scenarios.
Compared to a  setting in which a classifier was trained and testing using data from the same domain, we could reduce the performance gap from $10.5\,\%$ to $7.8\,\%$ with respect to the mean F1-score. 
For the two adaptation scenarios incorporating the domains with the most different class distributions according to the Jensen-Shannon-Divergence there was a rather high improvement of about $9\,\%$ in the mean F1-score due to DA. 
However, the resulting overall performance was still lower than the average performance after DA. 
We conclude that our proposed appearance adaptation method is able to achieve a high improvement even if the class distributions in the two domains are very different, but it cannot completely compensate for the domain gap. 
We assume that the comparably lower overall performance after DA to be caused by larger differences in the appearance of objects in the source and target domains.
We could show that our method outperforms existing methods for DA in RS applications, partly by a large margin. 
However, the remaining performance gap after DA (e.g., $7.8\,\%$ in mean F1-score; cf. above) indicates that there is still considerable room for improvement, even though a part of it may be attributed to systematic differences in the labelling policies applied for the generation of the test datasets used in our experiments, e.g. different rules to separate high and low vegetation. 


Having a look at the current state of domain adaptation for remote sensing applications, we think that appearance adaptation is currently the most stable way to achieve a positive transfer. 
However, the appearance adaptation using adversarial training remains susceptible to the choice of the architecture and the training strategy, but also to the distribution of the data. 
Although we have strong indications that our approach achieves semantically consistent transformation, this conclusion is mainly based on the visual evaluation and the relative improvements due to DA in the respective scenarios. 
For future works, it would be nice to evaluate semantic consistency in a more quantitative way.
One approach could be to evaluate the transformed images using a classifier that was trained in a supervised way in the target domain.
As our regularization loss for the discriminator resulted in a stabilization of the DA process, we  think that such a strategy could also be applied to adversarial representation matching, which we will investigate in our future work. 
We will also investigate if the method for appearance adaptation presented in this paper can be combined with approaches for representation transfer and even instance transfer to obtain a hybrid DA method that can further reduce the performance gap due to the lack of labelled data in the target domain. 
Lastly, we point out that the proposed entropy based parameter selection criterion, which was quite successful in the presented scenarios, can be transferred to many other settings, including supervised training. 
We think that particularly in scenarios where the amount of labelled training samples is limited, this approach for parameter selection is very useful, because it does only require a set of unlabelled samples.


\section*{Declaration of Competing Interest}
The authors declare that they have no known competing financial interests or personal relationships that could have appeared to influence the work reported in this paper.

\section*{Acknowledgements}
We thank the \textit{Landesamt f{\"u}r Vermessung und Geoinformation Schleswig Holstein} and the \textit{Landesamt f{\"u}r Geoinformation und Landesvermessung Niedersachsen (LGLN)} for providing the datasets. We also thank the International Society for Photogrammetry and Remote Sensing (ISPRS) for providing the data of the ISPRS labelling challenge. The Vaihingen dataset was provided by the German Society for Photogrammetry, Remote Sensing and Geoinformation (DGPF) \citep{Cramer2010}: \url{http://www.ifp.uni-stuttgart.de/dgpf/DKEP-Allg.html}.

\bibliography{2021_Wittich_DA_I2I}

\end{document}